%% file: longcat.tex
\documentclass{article}

\usepackage{longcat_style}
\usepackage{adjustbox}
\usepackage[utf8]{inputenc} %
\usepackage[T1]{fontenc}    %
\usepackage{newunicodechar}
\usepackage{hyperref}       %
\usepackage{xcolor}
\usepackage[normalem]{ulem} %
\hypersetup{
    colorlinks=true,      %
    linkcolor=blue,      %
    urlcolor=blue,       %
    citecolor=blue,      %
    linkbordercolor=blue, %
    urlbordercolor=blue,
    citebordercolor=blue,
    pdfborderstyle={/S/U/W 1}, %
}

\usepackage{float}
\usepackage{url}            %
\usepackage{booktabs}       %
\usepackage{amsfonts}       %
\usepackage{nicefrac}       %
\usepackage{microtype}      %
\usepackage{lipsum}		%
\usepackage{graphicx}
\usepackage{natbib}
\usepackage{doi}
\usepackage{amsmath}
\usepackage{amssymb} %
\usepackage{xspace}
\usepackage{enumitem}
\usepackage{multirow}
\usepackage{subcaption} 
\usepackage{makecell}
\usepackage{hyperref, cleveref}
\usepackage{pifont}
\usepackage[inkscapelatex=false]{svg}
\usepackage{subcaption}

\setlist[itemize]{leftmargin=*}
\setlist[enumerate]{leftmargin=*}
\setlist[description]{leftmargin=*}

\newcommand{\longcat}{LongCat-Flash\xspace}
\newcommand{\chip}{accelerators\xspace}

\newcommand{\kmax}{K}

\definecolor{midnightgreen}{rgb}{0.0, 0.29, 0.33}

\title{\longcat Technical Report}

\author{ Meituan LongCat Team \\
	\texttt{longcat-team@meituan.com} \\
}

\renewcommand{\headeright}{\raisebox{-0.2\height}{\includegraphics[height=1.8em]{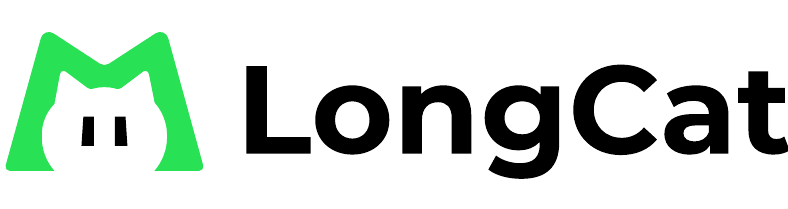}}}
\renewcommand{\shorttitle}{\longcat Technical Report}

\begin{document}
\maketitle

\begin{abstract}

We introduce \longcat, a 560-billion-parameter Mixture-of-Experts (MoE) language model designed for both computational efficiency and advanced agentic capabilities. Stemming from the need for scalable efficiency, \longcat adopts two novel designs: (a) \textit{Zero-computation Experts}, which enables dynamic computational budget allocation and activates 18.6B–31.3B (27B on average) per token depending on contextual demands, optimizing resource usage. (b) \textit{Shortcut-connected MoE}, which enlarges the computation-communication overlap window, demonstrating notable gains in inference efficiency and throughput compared to models of a comparable scale. We develop a comprehensive scaling framework for large models that combines hyperparameter transfer, model-growth initialization, a multi-pronged stability suite, and deterministic computation to achieve stable and reproducible training. Notably, leveraging the synergy among scalable architectural design and infrastructure efforts, we complete model training on more than 20 trillion tokens within 30 days, while achieving over 100 tokens per second (TPS) for inference at a cost of \$0.70 per million output tokens. To cultivate \longcat towards agentic intelligence, we conduct a large-scale pre-training on optimized mixtures, followed by targeted mid- and post-training on reasoning, code, and instructions, with further augmentation from synthetic data and tool use tasks. 
Comprehensive evaluations demonstrate that, as a non-thinking foundation model, \longcat delivers highly competitive performance among other leading models, with exceptional strengths in agentic tasks.
The model checkpoint of \longcat is open-sourced to foster community research.

\textbf{LongCat Chat}: \href{https://longcat.ai}{https://longcat.ai} \\
\textbf{Hugging Face}: \href{https://huggingface.co/meituan-longcat}{https://huggingface.co/meituan-longcat}\\
\textbf{GitHub}: \href{https://github.com/meituan-longcat}{https://github.com/meituan-longcat}

\end{abstract}

\begin{figure}[!htb]
    \centering
    \includegraphics[width=1\textwidth]{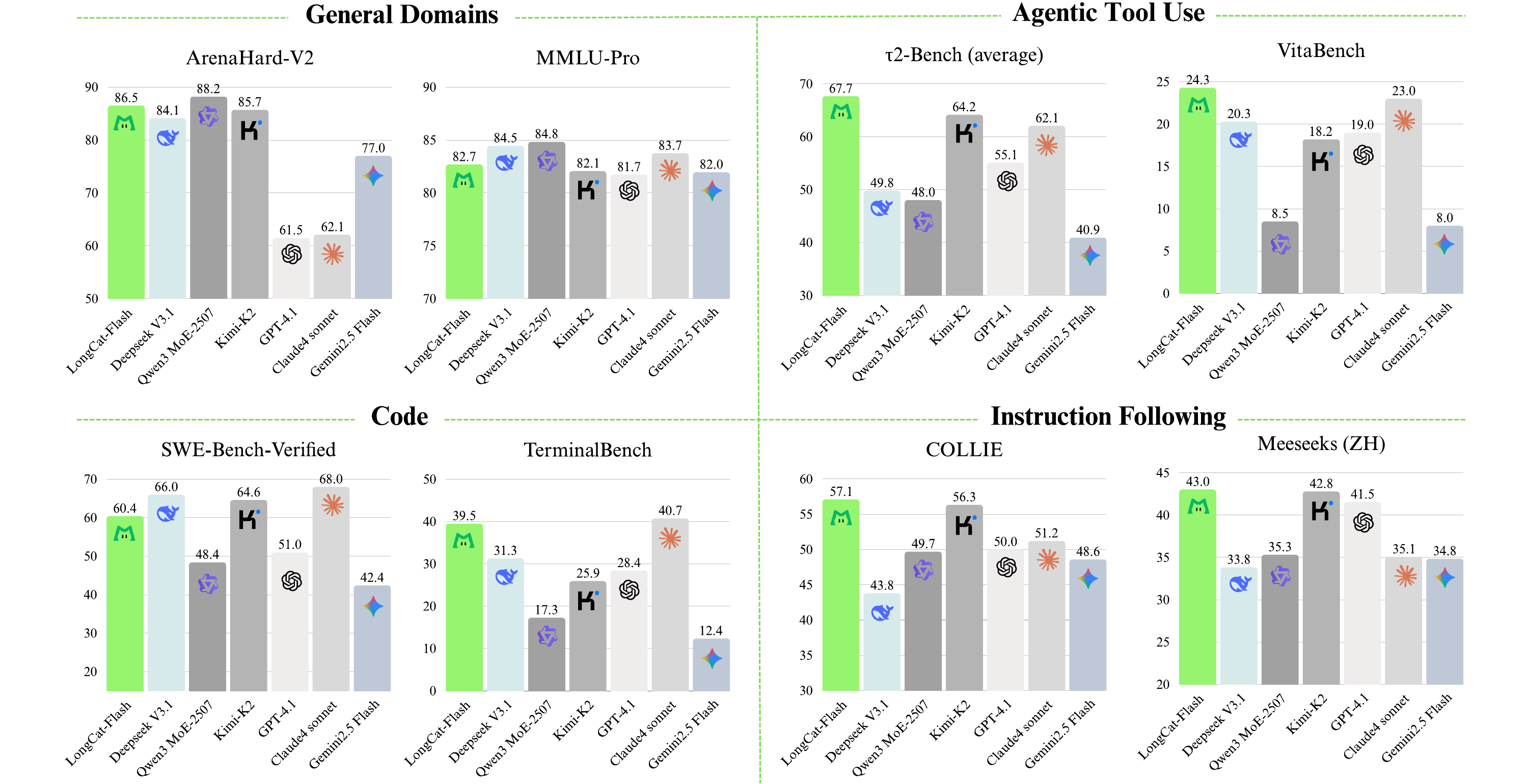}
    \caption{Benchmark performance of \longcat.}
\label{fig: benchmark_overview}
\end{figure}

\clearpage
\tableofcontents
\clearpage

\section{Introduction}

The rapid advancement of large language models (LLMs) such as DeepSeek-V3~\citep{deepseekai2025deepseekv3technicalreport}, Qwen 3~\citep{yang2025qwen3}, and Kimi-K2~\citep{team2025kimi} has demonstrated the effectiveness of scaling model size and computational resources. While some recent progress raises concerns about potential scaling slowdowns, we believe that algorithmic design, underlying system optimizations, and data strategy all play equally critical roles in further pushing the frontier of scalable intelligence. This requires innovations in both model architecture and training strategies to improve the cost-effectiveness of scaling, as well as a systematic data strategy to enhance the model's capability for solving real-world tasks.

In this work, we introduce \longcat, an efficient yet powerful Mixture-of-Experts (MoE) language model designed to advance the frontier of language model along two synergistic directions: \emph{computational efficiency} and \emph{agentic capability}. Trained on tens of thousands of \chip, \longcat combines architectural innovations with a sophisticated, multi-stage training methodology for scalable and intelligent models. Our contributions span both efficiency and agentic intelligence:

\begin{itemize}

\item \textbf{Scalable Architectural Design for Computational Efficiency} \longcat is designed and optimized under two key principles: efficient computation utilization, as well as efficient training and inference. Specifically, (1) As \textit{not all tokens are equal}, we introduce the \textit{zero-computation experts} mechanism in MoE blocks to allocate a dynamic computation budget to important tokens based on their significance, i.e., activating 18.6 to 31.3 billion parameters (out of 560 billion total) based on contextual demands. To ensure consistent computation load, we employ expert bias adjusted by a PID-controller, maintaining an average of $\sim$27 billion activated parameters per token. (2) As communication overhead becomes a bottleneck during MoE model scaling, we incorporate the \textit{Shortcut-connected MoE (ScMoE)} \citep{cai2024shortcut} design to expand the computation-communication overlap window. Combined with customized infrastructure optimizations, this design  enables training at a massive scale of over tens of thousands \chip and inference with high throughput and low latency.

\item \textbf{Effective Model Scaling Strategy} Effectively and efficiently scaling model size remains a key challenge in strategy design. To this end, we develop a comprehensive stability-and-scaling framework for robustly training large-scale models: (1) We successfully apply a hyperparameter transfer strategy to such a large model, predicting optimal hyperparameter configurations by leveraging results from smaller proxy models with theoretical guarantees. (2) We initialize the model using a model-growth mechanism based on a refined half-scale checkpoint, achieving improved performance compared to conventional initialization methods. (3) A multi-pronged stability suite incorporates principled router-gradient balancing, a hidden z-loss to suppress massive activations, and fine-tuned optimizer configurations. (4) To enhance the reliability of large-scale cluster training, we introduce deterministic computation. This guarantees the exact reproducibility of experiments and enables the detection of SDC (Silent Data Corruption) during the training process. These interventions ensure that \longcat’s training remains stable, with no irrecoverable loss spikes.

\item \textbf{Multi-Stage Training Pipeline for Agentic Capability} 
Through a meticulously designed pipeline, \longcat is endowed with advanced agentic behaviors. 
Initial efforts focus on constructing a more suitable base model for agentic post-training, where we design a two-stage pretraining data fusion strategy to concentrate reasoning-intensive domain data. During mid-training, we enhance reasoning and coding capabilities while extending the context length to 128k to meet agentic post-training requirements.
Building on this advanced base model, we proceed with a multi-stage post-training. Recognizing the scarcity of high-quality, high-difficulty training problems for agentic tasks, we design a multi-agent synthesis framework that defines task difficulty across three axes, i.e., information processing, tool-set complexity, and user interaction—using specialized controllers to generate complex tasks requiring iterative reasoning and environmental interaction.

\end{itemize}

Overall, benefiting from our synergy among scalable architectural design, training strategies, and infrastructure efforts, \longcat achieves both high training throughput and low inference latency.
Notably, we complete the pre-training of our 560B model over 20T tokens within 30 days and achieve 98.48\% time availability without manual intervention for fault resolution.
During inference, large-scale deployment efficiency exceeds 100 tokens per second (TPS) on H800, with a cost of \$0.7 per million output tokens, demonstrating remarkable performance compared to models with similar size.

We evaluate the base and instruction-tuned versions of \longcat across diverse benchmarks, with an overview summarized in Figure~\ref{fig: benchmark_overview}. As a non-thinking model, \longcat achieves performance comparable to state-of-the-art non-thinking models, including DeepSeek-V3.1~\citep{deepseekai2025deepseekv3technicalreport} and Kimi-K2~\citep{team2025kimi}, while using fewer parameters and offering faster inference speed. Specifically, \longcat scores 86.5 on ArenaHard-V2, 39.5 on TerminalBench, and 67.7 on $\tau^2$-Bench, demonstrating robust capabilities in general domains, coding, and agentic tool use. To mitigate potential contamination from existing open-source benchmarks and enhance evaluation confidence, we meticulously constructed two new benchmarks: Meeseeks~\citep{meeseeks} and VitaBench. Meeseeks simulates realistic human-LLM interactions through an iterative feedback framework to evaluate multi-turn instruction-following ability, where \longcat achieves scores on par with frontier LLMs. VitaBench leverages real-world business scenarios to access models’ proficiency in addressing complex real-world tasks, where \longcat delivers superior performance than other LLMs.

In the remainder of this report, we first detail the architecture and innovations in \longcat. Then, we describe the pre-training and post-training processes, including our training strategies, data construction methods, and evaluation results. Finally, we discuss the challenges and solutions in training \longcat, along with optimized inference and deployment methods that leverage its unique architecture.

\section{Architecture}

\begin{figure}
    \centering
    \includegraphics[width=0.8\linewidth]{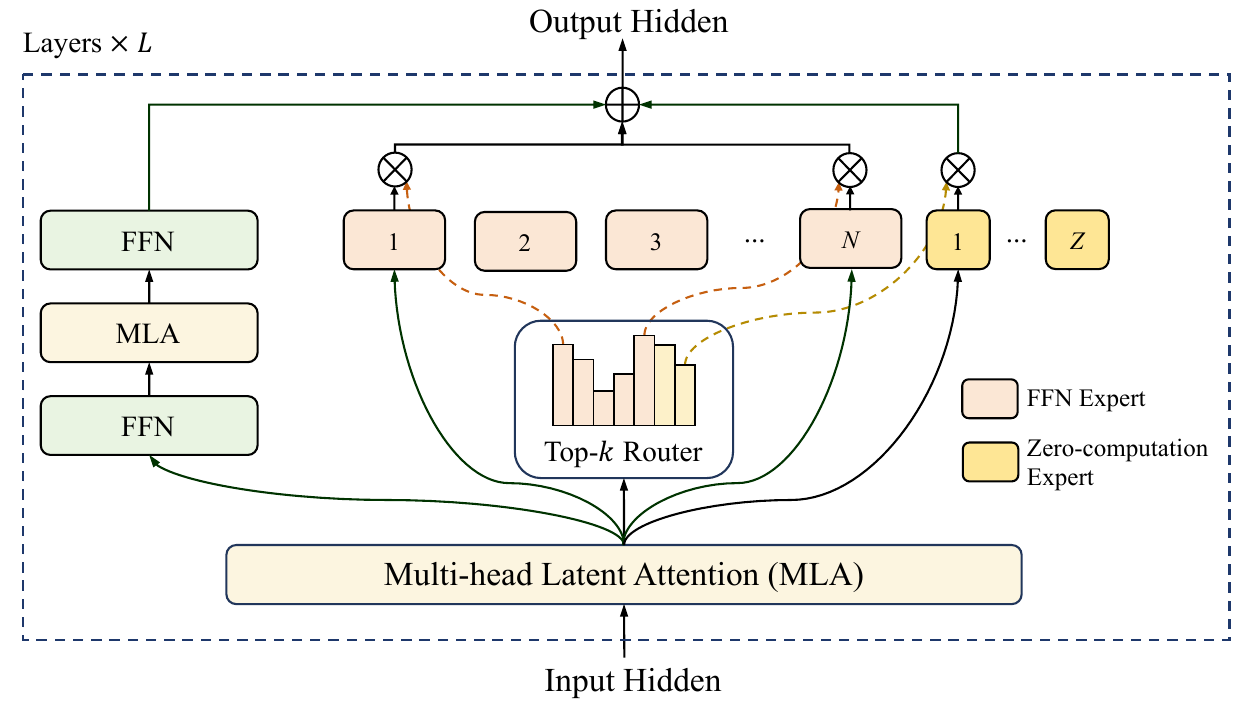}
    \caption{The architecture adopted in \longcat. Each layer employs Shortcut-connected Mixture of Experts (ScMoE) with zero-computation experts. ScMoE significantly expands the computation-communication window to boost training and inference efficiency. The zero-computation experts enable dynamic computation based on contextual importance, improving the efficiency of computational resource utilization.}
    \label{fig:longcat-moe}
\end{figure}

\longcat adopts a novel MoE architecture with two key innovations (Figure~\ref{fig:longcat-moe}): 
(1) The MoE block incorporates zero-computation experts \citep{jin2024moe++} to enable dynamic computation, allowing tokens to consume variable computational resources based on their contextual significance. Furthermore, the average computational load is regulated through an adaptive expert bias.
(2) Each layer integrates two Multi-head Latent Attention (MLA)  block \citep{liu2024deepseek} and multiple heterogeneous Feed-Forward Network (FFN) blocks. A \emph{shortcut} connection from the first MLA output directly to the MoE block \citep{cai2024shortcut} is employed.  To further enhance performance, we refine both the MLA and fine-grained FFN experts via variance alignment. The following subsections will detail each of these components.

\subsection{Zero-Computation Experts}
Next-token prediction exhibits inherent computational heterogeneity. Difficult tokens may demand more resources for accurate prediction, while easy tokens require negligible computation. This phenomenon is also empirically evidenced by speculative decoding, where small draft models reliably predict the outputs of large models for most easy tokens \citep{leviathan2023fast}. 

Motivated by this, \longcat presents a dynamical computational resource allocation mechanism by activating a variable number of FFN experts per token through zero-computation experts \citep{jin2024moe++, zeng-etal-2024-adamoe}, enabling a more reasonable allocation of computations according to contextual significance.   
Specifically, \longcat expands its expert pool with $Z$ zero-computation experts  in addition to $N$ standard FFN experts. Zero-computation experts simply return the input $x_t$ as their output, thereby introducing no additional computational cost. Let $x_t$ be the MoE input of the $t$-th token, the MoE module in \longcat can be formulated as follows:
\begin{equation}
\begin{aligned}
    \text{MoE}(x_t) &= \sum_{i=1}^{N+Z} g_i \, E_i(x_t), \\[6pt]
    g_i &= 
    \begin{cases} 
        R(x_t)_i, & \text{if } R(x_t)_i \in \text{TopK}\!\big(R(x_t)_i + b_i \;\big|\; 1 \leq i \leq N+Z, \kmax\big), \\[6pt]
        0, & \text{otherwise}, 
    \end{cases} \\[10pt]
    E_i(x_t) &= 
    \begin{cases} 
        \text{FFN}_i(x_t), & \text{if } 1 \leq i \leq N, \\[6pt]
        x_t, & \text{if } N < i \leq N+Z,
    \end{cases}
\end{aligned}
\end{equation}

\noindent
where $R$ denotes the softmax router, $b_i$ is the expert bias corresponding to the $i$-th expert, and $\kmax$ denotes the number of experts selected per token.

\begin{figure}[t]
    \centering
    \begin{subfigure}[b]{0.33\linewidth}
        \centering
        \includegraphics[width=\linewidth]{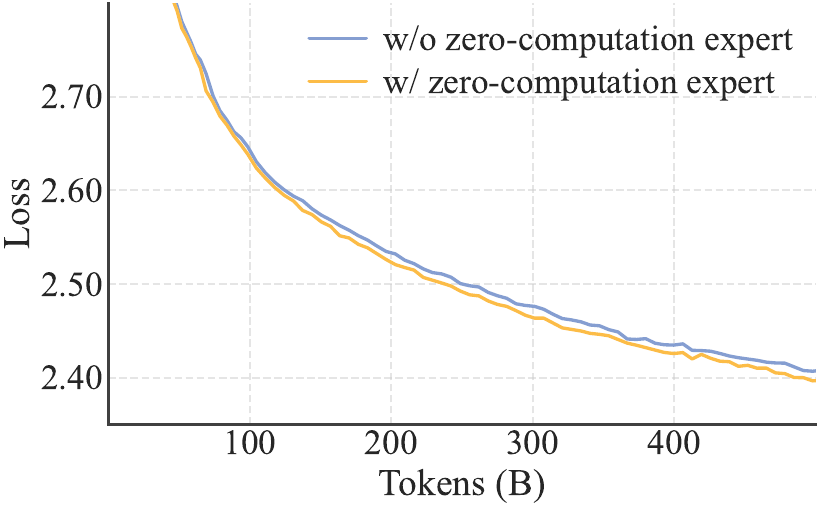}
        \caption{}
        \label{fig:valid-loss-mmlu}
    \end{subfigure}
    \hfill
    \begin{subfigure}[b]{0.33\linewidth}
        \centering
        \includegraphics[width=\linewidth]{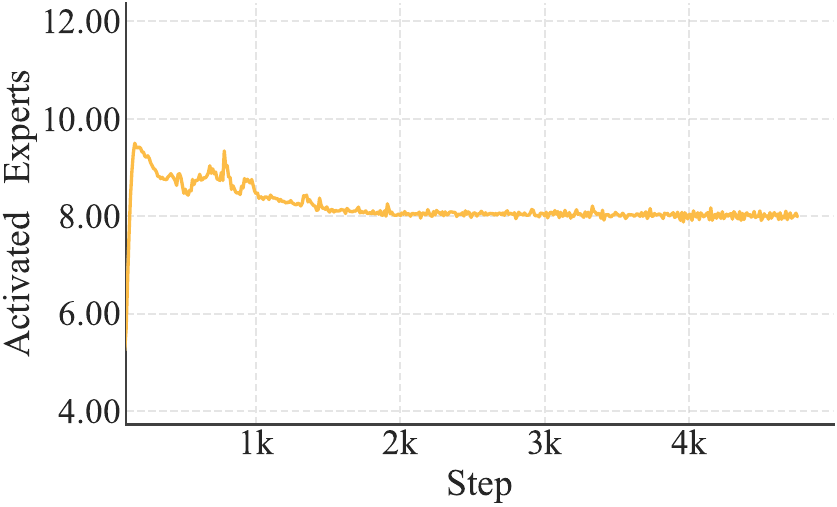}
        \caption{}
        \label{fig:avg-topk}
    \end{subfigure}
    \hfill
    \begin{subfigure}[b]{0.33\linewidth}
        \centering
        \includegraphics[width=\linewidth]{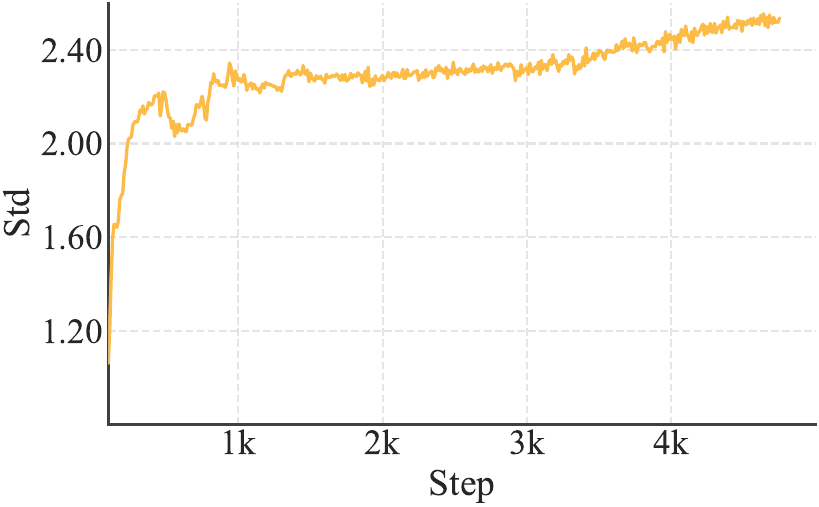}
        \caption{}
        \label{fig:select-std}
    \end{subfigure}
    \caption{\textbf{(a)} Validation loss curve comparing models with/without zero-computation experts under matched computation budgets. The baseline (top-k=8, blue) activates fixed 6B parameters per token, while the zero-expert variant (top-k=12, orange) dynamically activates 4.2B-7.0B parameters but maintains 8 FFN experts expectation (with fluctuation less than 1\%). The consistent loss reduction demonstrates the efficacy of zero-computation experts. \textbf{(b)} The average number of activated FFN experts during \longcat training. The average number remains closely around 8, corresponding to expected 27B activated parameters. \textbf{(c)} The standard deviation of activated FFN experts grows to 3, indicating substantial variability in activated parameters across different tokens.}
    \label{fig:valid-loss-5d6b}
\end{figure}

The router assigns each token to $\kmax$ experts, where the number of activated FFN experts varies per token based on contextual importance. Through this adaptive allocation mechanism, the model learns to dynamically allocate more computational resources to tokens with higher contextual importance,  thus achieving superior performance under the same computational capacity as illustrated in Figure~\ref{fig:valid-loss-mmlu}.

\subsubsection{Computational Budget Control}
To incentivize the model to learn context-dependent computation allocation, fine-grained control over the average selection ratio of zero-computation experts is essential. Without explicit constraints, the model tends to under-utilize zero-computation experts, leading to inefficient resource usage.

We accomplish this by refining the expert bias mechanism from the aux-loss-free strategy \citep{wang2024auxiliary}, introducing an expert-specific bias term that dynamically adjusts routing scores based on recent expert utilization, while remaining decoupled from the language model (LM) training objective. 
For the expert bias $b_i$ corresponding to the $i$-th expert, it is updated each step with the incremental computed as:
\begin{equation}
\label{eq:bias}
    \Delta b_i =
    \begin{cases}
        \mu \left( \dfrac{K_{e}}{K} \cdot \dfrac{1}{N} \;-\; \dfrac{T_i}{KT_{\mathrm{all}}} \right), & \text{if } 1 \leq i \leq N, \\[1.2ex]
        0, & \text{if } N < i \leq N+Z,  \\
    \end{cases}
\end{equation}
where $\mu$ denotes the bias adaptation rate, $T_{\mathrm{all}}$ denotes the  number of tokens in a global batch, $T_i$ denotes the number of tokens routed to the $i$-th expert, $K_{e}$ denotes the expected number of activated FFN experts, which is smaller than $K$.

The proposed update rule employs a PID controller (proportional-integral-derivative) from control theory \citep{control_book}, ensuring that the token allocation for the $i$-th expert converges to its target proportion. Compared to a fixed bias increment~\citep{wang2024auxiliary}, this mechanism improves the robustness of the softmax router's probability distribution as the number of experts scales. Notably, we exclude zero-computation experts from bias updates, as their identity nature only requires a global constraint, which is automatically satisfied when all FFN experts achieve their expected token proportions. Empirically, large batch sizes and a decay schedule for $\mu$ improve the stability of budget control, while small batch sizes may require reduced update frequency.

During pre-training, we tracked the average number and standard deviation of activated experts (Figure~\ref{fig:avg-topk} and \ref{fig:select-std}). The results show that after approximate 20B tokens of adjustment, the average expert number in all layers converged to the expected value, with fluctuations less than 1\%.
However, the standard deviation persisted at a relatively high level, indicating that the model allocates substantially divergent computational resources across tokens.

For detailed statistics and case studies of dynamic routing, please refer to Appendix~\ref{appendix:pretrain_dynamics}.

\subsubsection{Load Balance Control}
Efficient MoE training requires robust load balancing among FFN experts. While Eq.~\eqref{eq:bias} enforces balance at the corpus level, a device-level load balance loss \citep{deepseekai2025deepseekv3technicalreport} to further prevent extreme sequence-level imbalance among EP groups is introduced. We make necessary efforts to accommodate the zero-computation experts. Specifically, assuming that all $N$ FFN experts are divided into $D$ groups, each group containing $G=\frac{N}{D}$ experts, the loss can be expressed as:
\begin{align}
    \mathcal{L}_{\text{LB}} &= \alpha \sum_{j=1}^{D+1} f_j P_j, \label{eq:loss_lb}\\
    P_j &= \frac{1}{T} \sum_{i \in \text{Group}_j}\sum_{t=1}^T R(x_t)_i, \label{eq:pj}\\
    f_j &= \begin{cases}
        \dfrac{D}{K_e T}\displaystyle\sum_{t=1}^T \mathbb{I}(\text{token } t \text{ selects Group}_j), & \text{if } 1 \leq j \leq D, \\[3mm]
        \dfrac{1}{(K-K_e)T}\displaystyle\sum_{t=1}^T \mathbb{I}(\text{token } t \text{ selects zero-computation experts}), & \text{if } j = D+1,
    \end{cases} \label{eq:fj}
\end{align}
where $\alpha$ is the balance factor, $T$ is the number of tokens in a micro batch, and $\mathbb{I}$ denotes the indicator function. In the loss, we assign all zero-computation experts to an additional group and average the frequency in each group. By adjusting the coefficient of $f_j$, we ensure that the ratio of FFN experts to zero-computation experts would approach $\frac{K_e}{\kmax-K_e}$ when the loss converges.
\begin{figure}[t!] %
    \centering
    
    \begin{subfigure}[b]{0.33\textwidth}
        \centering
        \includegraphics[width=\textwidth]{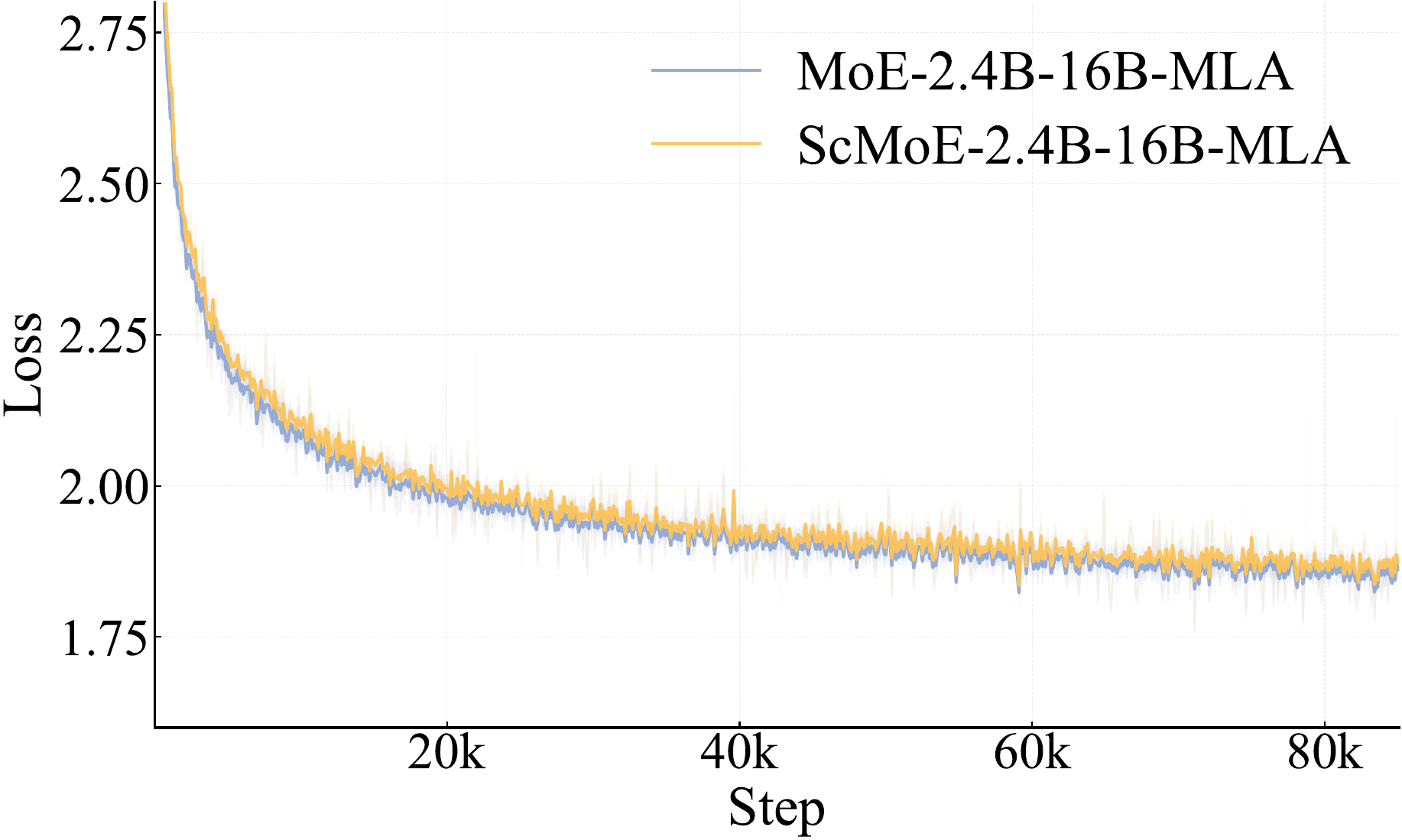} %
        \subcaption{}
        \label{fig:loss-2.5b}
    \end{subfigure}
    \hfill %
    \begin{subfigure}[b]{0.33\textwidth}
        \centering
        \includegraphics[width=\textwidth]{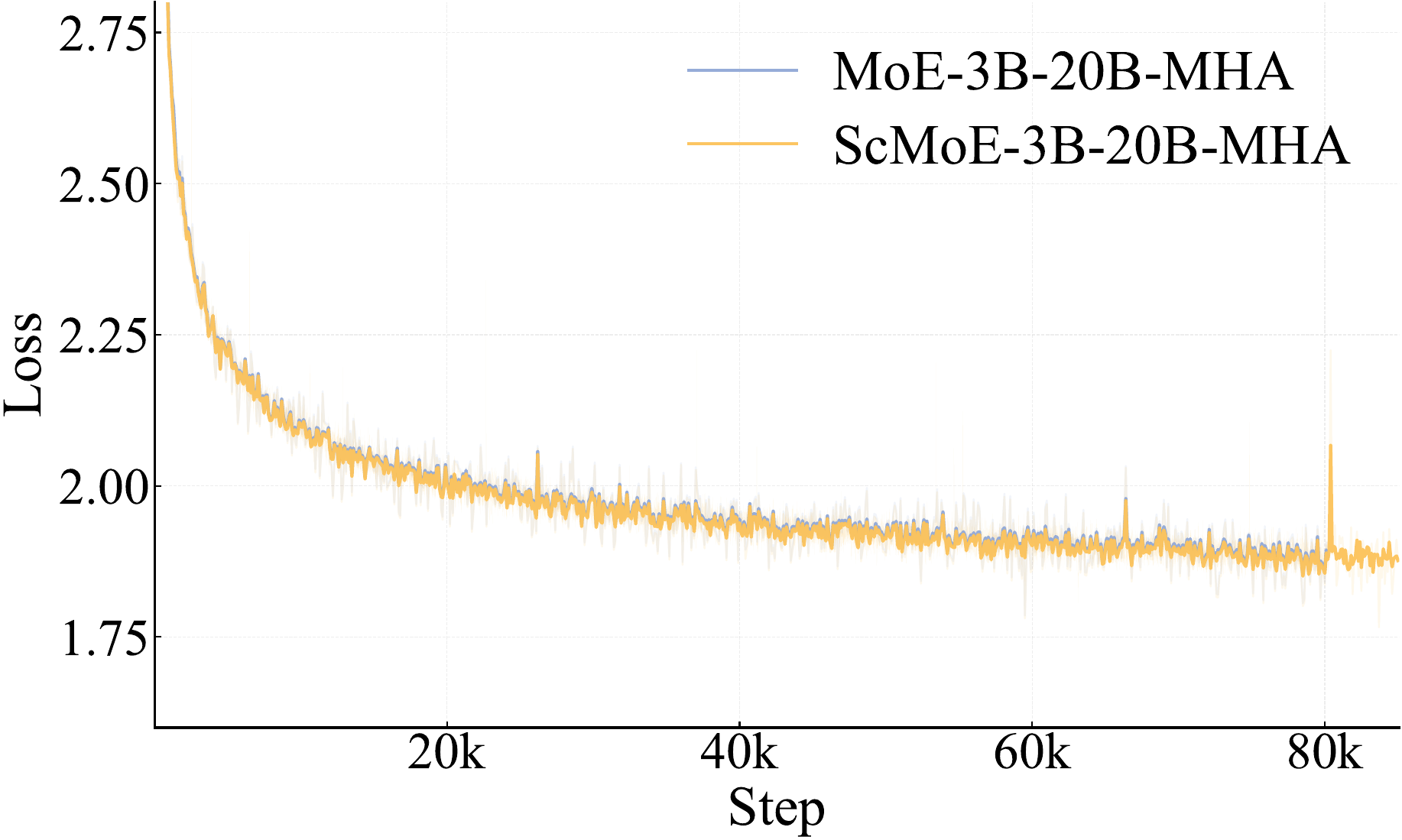} %
        \subcaption{}
        \label{fig:loss-3b}
    \end{subfigure}
    \hfill %
    \begin{subfigure}[b]{0.33\textwidth}
        \centering
        \includegraphics[width=\textwidth]{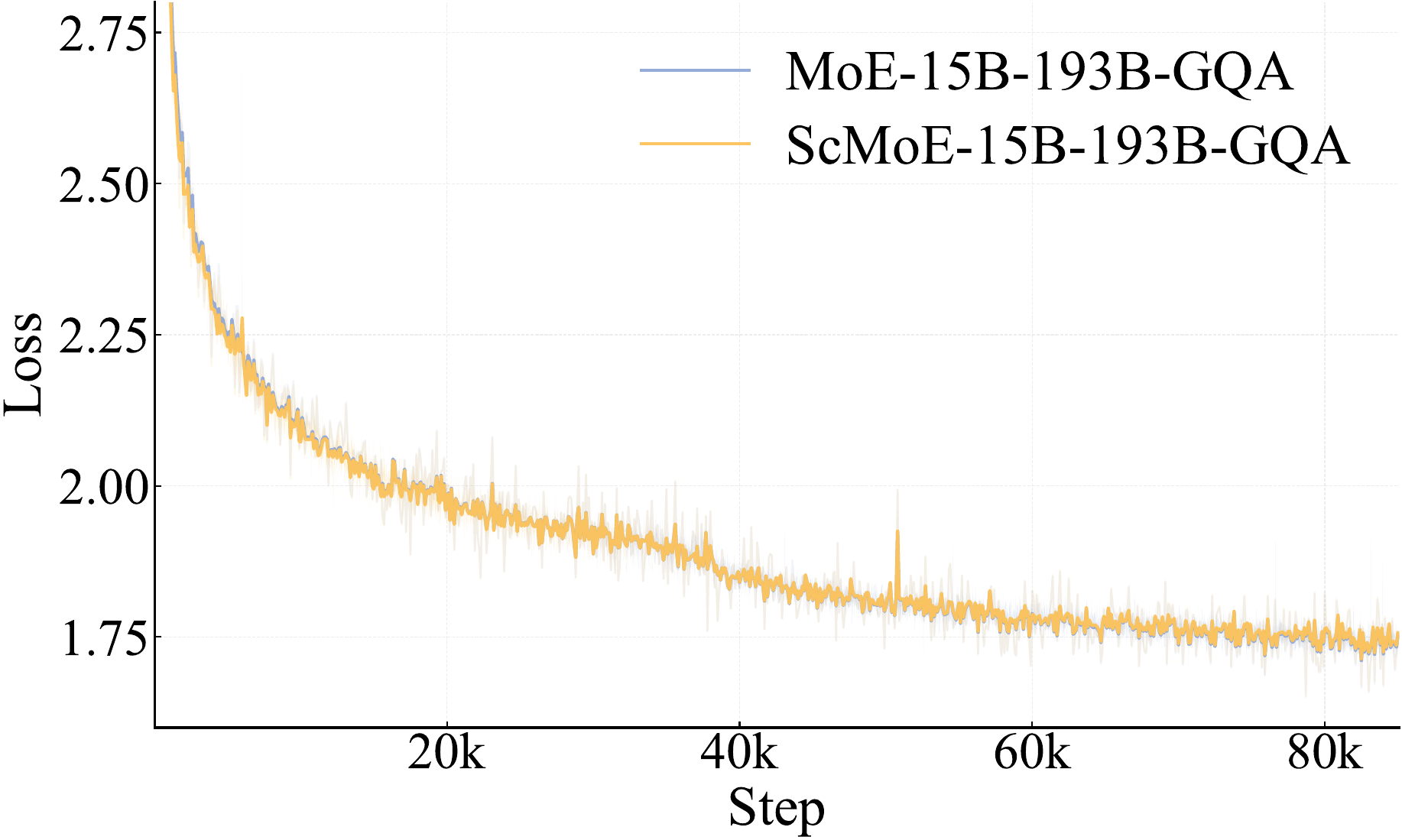} %
        \subcaption{}
        \label{fig:loss-15b}
    \end{subfigure}

    \caption{
        Training loss curves comparing baseline models (without ScMoE) against their ScMoE-enhanced counterparts across four different model configurations. In all experiments—\textbf{(a)} 2.4B-16B with MLA, \textbf{(b)} 3B-20B with MHA, and \textbf{(c)} 15B-193B with GQA—the loss curves are virtually indistinguishable. This provides robust evidence that the ScMoE optimization is quality-neutral, and its benefits are orthogonal to both model scale and the specific attention architecture used.
    }
    \label{fig:loss-curves-scmoe} %
\end{figure}

\subsection{Shortcut-Connected MoE}
Our initial architecture employs an interleaved topology of MoE and dense FFN blocks. This design has been extensively validated through empirical studies, demonstrating performance comparable to leading shared-expert models \citep{deepspeed, liu2024deepseek}. However, the efficiency of large-scale MoE models is largely constrained by communication overhead.  In the conventional execution paradigm, expert parallelism imposes a sequential workflow: an  collective operation must first route tokens to their designated experts before computation can begin. This communication latency becomes a bottleneck, leading to device underutilization and limiting overall system throughput.

While shared-expert architectures attempt to mitigate this by overlapping communication with a single expert's computation, their efficiency is limited by the small computational window of that one expert. We overcome this limitation by employing the Shortcut-connected MoE (ScMoE) architecture \citep{cai2024shortcut}. ScMoE introduces a cross-layer shortcut that reorders the execution pipeline. This key innovation allows the dense FFN from the preceding block to execute in parallel with the dispatch/combine communication of the current MoE layer, creating a more substantial overlap window than shared-expert designs. Furthermore, the architecture design choice is verified by the following key findings.

First, ScMoE structure does not compromise model quality. As shown in Figure~\ref{fig:loss-curves-scmoe}, the training loss curves of our architecture and the baseline without ScMoE are nearly identical, confirming this reordered execution does not impair model performance. Consistent results are observed across multiple settings, including a 2.4B-16B MoE model with MLA, a 3B-20B model with MHA~\citep{vaswani2017attention}, and 15B-193B models with GQA~\citep{ainslie2023gqa}. Importantly, these findings demonstrate that the stability and benefits of ScMoE are orthogonal to the choice of attention mechanism.

Second, the ScMoE architecture delivers substantial system-level efficiency gains for both training and inference.

\textbf{For Large-Scale Training}: The expanded overlap window allows the computation of the preceding block to be fully parallel with its dispatch and combine communication phases in the MoE layer, achieved by partitioning operations into fine-grained chunks along the token dimension.

\textbf{For Efficient Inference}: ScMoE enables a \textit{Single Batch Overlap} pipeline, reducing the theoretical Time-Per-Output-Token (TPOT) by nearly 50\% compared to leading models such as DeepSeek-V3. Moreover, it allows for the concurrent execution of distinct communication patterns: intra-node Tensor Parallelism communication (via NVLink) on the dense FFN can be fully overlapped with inter-node Expert Parallelism communication (via RDMA), thereby maximizing total network utilization.

In summary, ScMoE delivers substantial performance gains without sacrificing model quality. These efficiency gains are not achieved through trade-offs but are the direct outcome of a rigorously validated, quality-neutral architectural innovation.

\subsection{Variance Alignment Design for Scalability}
Architectural designs that excel at small scales may become suboptimal as models are scaled up, and vice versa, rendering initial design choices unreliable. Through extensive experimentation and theoretical analysis, we identify \textit{variance misalignment} in specific modules as a key factor contributing to this discrepancy, which can lead to instability and degraded performance during scaling. To address this challenge, we propose variance alignment techniques for both MLA and MoE blocks.

\subsubsection{Scale-Correction for MLA}
\label{sssec:mla_variance_alignment}
\longcat employs a modified Multi-head Latent Attention (MLA) mechanism~\citep{liu2024deepseek}, which incorporates scale-correction factors $\alpha_q$ and $\alpha_{kv}$ to address the variance imbalances inherent in asymmetric low-rank factorization. Our full mathematical formulation, which integrates these correction factors, is given as follows:
\begin{equation}
\begin{aligned}
    &c_t^Q   = \boxed{\alpha_q}\, W^{DQ} h_t \in \mathbb{R}^{d_q}, 
    &\quad &c_t^{KV} = \boxed{\alpha_{kv}}\, W^{DKV} h_t \in \mathbb{R}^{d_{kv}}, \\[0.8ex]
    &q_{t,1:n_h}^C = W^{UQ} c_t^Q, 
    &\quad &k_{t,1:n_h}^C = W^{UK} c_t^{KV}, 
    &\quad v_{t,1:n_h}   &= W^{UV} c_t^{KV}, \\[0.8ex]
    &q_{t,1:n_h}^R = \mathrm{RoPE}(W^{QR} c_t^Q), 
    &\quad &k_t^R     = \mathrm{RoPE}(W^{KR} h_t), \\[0.8ex]
    &q_{t,i}   = \big[\, q_{t,i}^C \,;\, q_{t,i}^R \,\big], 
    &\quad &k_{t,i}   = \big[\, k_{t,i}^C \,;\, k_t^R \,\big], \\[0.8ex]
    &o_{t,i} = \mathrm{Attention}\!\left(q_{t,i},\, k_{1:t,i},\, v_{1:t,i}\right), 
    &\quad &u_t = W^O \big[\, o_{t,1} \,;\, o_{t,2} \,;\, \dots \,;\, o_{t,n_h} \,\big],
\end{aligned}
\end{equation}

where $h_t \in \mathbb{R}^{d_{\text{model}}}$ is the attention input of the $t$-th token, and $n_h$ is the number of heads.

The introduction of $\alpha_q$ and $\alpha_{kv}$ address a fundamental variance mismatch among query/key vector components. At initialization, their variances are proportional to their source dimensions: $\sigma^2(q_t^C), \sigma^2(q_t^R) \propto d_q$ and $\sigma^2(k_t^C) \propto d_{kv}$. In contrast, the rotary key component $k_t^R$ has a variance proportional to the full model dimension: $\sigma^2(k_t^R) \propto d_{\text{model}}$. This dimensional disparity  causes unstable attention scores at initialization when $d_q$, $d_{kv}$, and $d_{\text{model}}$ are varied, resulting in degraded and unpredictable performance during model scaling.

\begin{figure}[t]
    \centering
    \begin{subfigure}[b]{0.4\linewidth}
        \includegraphics[width=\linewidth]{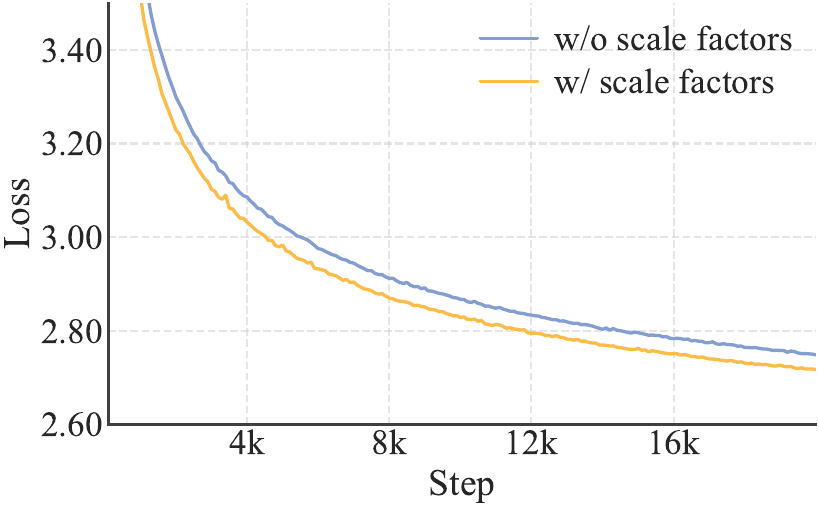}
        \caption{}
        \label{fig:mla_scale}
    \end{subfigure}
    \hspace{0.1\linewidth}
    \begin{subfigure}[b]{0.4\linewidth}
        \includegraphics[width=\linewidth]{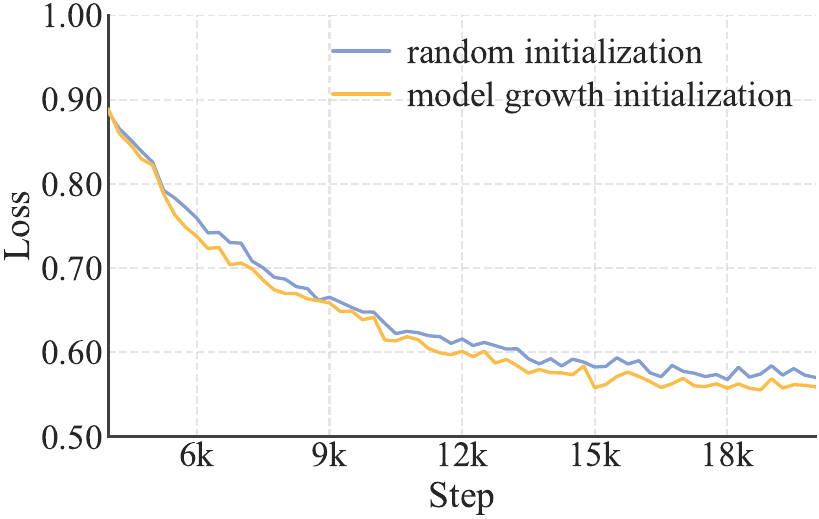}
        \caption{}
        \label{fig:model_grrowth_fig2}
    \end{subfigure}
    \caption{\textbf{(a)} Incorporating the scale-correction factor on MLA showing improved convergence (lower loss) on a 1B activated MOE model. \textbf{(b)} Validataion loss curve of a 6B activated MoE model in model growth experiments.}
    \label{fig:model_growth_val_multi_fig}
\end{figure}

Our solution is to rescale the low-rank path components to align their final variance with a reference scale, and we use the full model dimension as a reference. This is achieved by defining the scaling factors as:
\begin{equation}
    \alpha_q = \sqrt{\frac{d_{\text{model}}}{d_q}} \quad \text{and} \quad \alpha_{kv} = \sqrt{\frac{d_{\text{model}}}{d_{kv}}}.
\end{equation}
This scale-invariant correction neutralizes the variance mismatch, ensuring they are well-conditioned for the attention computation. Our experiments reveal that this method improves the model performance, as shown in Figure~\ref{fig:mla_scale}.

\subsubsection{Variance Compensation for Experts Initialization}
\longcat adopts the fine-grained expert strategy from DeepSeek-MoE \citep{liu2024deepseek}, which segments each expert into $m$ finer-grained ones to enhance combinatorial flexibility and knowledge specialization. However, we observe that the performance of this design is sensitive to other architectural choices (e.g., expert numbers, top-k, $m$). To address this, we propose a variance compensation mechanism that counteracts the initialization variance reduction caused by expert segmentation. The mechanism applies a scaling factor $\gamma$ to the aggregated output of the experts, formulated as:
\begin{equation}
    \text{MoE}(x_t) = \gamma \left( \sum_{i=1}^{mN} g_i \cdot E_i(x_t) \right),
    \label{eq:compensated_moe}
\end{equation}
where $g_i$ is the router output over $mN$ fine-grained experts and $N$ represents the total number of experts before segmentation.

The scaling factor $\gamma$ in Eq.~\eqref{eq:compensated_moe} is derived by quantifying two primary sources of variance reduction:
\begin{enumerate}
    \item Gating Dilution: Decomposing each original $N$ experts into $m$ finer-grained experts expands the total expert counts to $mN$. This expansion forces the softmax gate to distribute its probability mass across a larger expert pool, proportionally reducing the magnitude of individual gating values $g_i$. Consequently, the output variance is reduced approximately by a factor of $m$.
    \item Dimensional Reduction: The intermediate hidden dimension of each fine-grained expert ($d_{\text{expert\_inter}}$) is reduced by a factor of $m$. Assuming uniform parameter initialization, the output variance of a single expert also decreases by a factor of $m$.
\end{enumerate}

To preserve the MoE layer’s output variance at initialization (matching the pre-segmentation baseline), $\gamma$ must compensate for both effects. The combined variance compensation factor is thus $\gamma = \sqrt{m \cdot m}= m$.

\subsection{Model Information}

\paragraph{Tokenizer} \longcat employs byte-pair encoding (BPE) \citep{shibata1999bpe,sennrich2015neural} for tokenization. Our tokenizer is trained on a comprehensive multilingual corpus spanning web pages, books, source code, etc, ensuring robust cross-domain performance. While inheriting GPT-4's pre-tokenization framework, we introduce the following modifications: (1) Enhanced CJK character segmentation for improved Chinese text handling, and (2) Independent digit tokenization to boost mathematical capabilities. The vocabulary size is optimized at 131,072 tokens, striking an effective balance between computational efficiency and linguistic coverage.

\paragraph{Multi-Token Prediction} To enhance inference efficiency, we integrate Multi-Token Prediction (MTP) \citep{gloeckle2024better,deepseekai2025deepseekv3technicalreport} as an auxiliary training objective. For optimal inference performance, we employ a single dense layer rather than a MoE layer as the MTP head. Empirical observations reveal rapid convergence of MTP loss, prompting us to strategically introduce MTP training in the middle of the training pipeline to balance model performance with prediction accuracy. The MTP head achieves >90\% acceptance rate in evaluations (Table~\ref{table:mtp_speedup}). 

\paragraph{Model Configurations} \longcat consists of 28 layers (excluding the MTP layer) with a 6144-dimensional hidden state. Each MLA block uses 64 attention heads with per-head dimension 128 for balanced performance-efficiency tradeoff. Following DeepSeek-V3 \citep{liu2024deepseek}, we set the KV compression dimension to 512, and the query compression dimension to 1536. The FFNs in the dense path employ 12288 intermediate dimensions, while each FFN expert uses 2048 dimensions. The scaling factors in MLA blocks and FFN blocks follow the methodology in Section~\ref{sssec:mla_variance_alignment}. Each layer contains 512 FFN experts and 256 zero-computation experts,  with exactly 12 experts activated per token (selected from both types). \longcat has 560B total parameters, activating between 18.6B and 31.3B parameters per token depending on context, with an average activation of approximately 27B parameters.

\section{Pre-Training}

The pre-training of \longcat follows a three-stage curriculum: (1) We train the model on approximately 20 trillion tokens with 8192 sequence length to establish a robust base model. (2) Reasoning and coding capabilities are further enhanced using trillions of data. (3) The context length is extended to 128k through training on long context corpora. Each stage implements tailored data strategies accompanied by rigorous decontamination procedures to prevent test set leakage.

To optimize scalability, we introduce hyperparameter transfer and model growth strategies, significantly improving performance as model size increases. Given the inherent instability challenges in large-scale training, we identify and implement multiple effective techniques to enhance training stability.

\subsection{Training Strategy}

\subsubsection{Hyperparameter Transfer}
\longcat employs a hyperparameter transfer strategy based on width scaling \citep{everett2024scaling} to efficiently train large-scale models. The methodology involves: (1) identifying optimal hyperparameters on a smaller proxy model, and (2) transferring these configurations to the target model through theoretically-motivated scaling rules.

The transfer mechanism centers on the width scaling factor $s = n_{\text{target}} / n_{\text{proxy}}$, where $n$ is the model's hidden dimension. We specifically adopt the ``Adam LR Full Align'' rules for Standard Parameterization. These rules specify how to adapt the proxy model's optimal initialization variance ($\sigma^2$) and learning rate ($\eta$) for the target architecture. The practical transfer rules are summarized in Table~\ref{tab:practical_transfer}. 

\begin{table}[h!]
\centering
\caption{Practical hyperparameter transfer rules and their underlying scaling exponents, derived from the Adam LR Full Align principle for Standard Parameterization \citep{everett2024scaling}. Here, $s$ is the width scaling factor $n_{\text{target}} / n_{\text{proxy}}$.}

\label{tab:practical_transfer}
\begin{tabular}{lcl}
\toprule
\textbf{Layer \& Parameter}  & \textbf{Target Model Setting} \\ 
\midrule
Embedding (Init Var, $\sigma^2$) & $\sigma^2_{\text{target}} = \sigma^2_{\text{proxy}}$ \\
Embedding (Learning Rate, $\eta$) & $\eta_{\text{target}} = \eta_{\text{proxy}}$ \\ 
\midrule
Hidden/Unembedding (Init Var, $\sigma^2$) & $\sigma^2_{\text{target}} = \sigma^2_{\text{proxy}} / s$ \\
Hidden/Unembedding (Learning Rate, $\eta$)  & $\eta_{\text{target}} = \eta_{\text{proxy}} / s$ \\ 
\bottomrule
\end{tabular}
\end{table}

Following this methodology, our training involves the following steps:
\begin{enumerate}
    \item We set the width scaling factor $s$ to 8 based on a trade-off analysis between computational efficiency and transfer performance. The proxy model is configured with a width of 768.
    \item We then perform a comprehensive hyperparameter search on the proxy model to identify the optimal layer-specific initializaton variances ($\sigma^2_{\text{proxy}}$) and learning rates ($\eta_{\text{proxy}}$).
    \item The optimal hyperparameters from the proxy model were transferred to the target model following the rules detailed in Table~\ref{tab:practical_transfer}. All other architectural attributes (depth, sparsity, and batch size) remain invariant during this transfer process.
\end{enumerate} 

We conducted comprehensive experiments to validate the effectiveness of this approach. The results demonstrate that this method significantly reduces computational costs when identifying optimal hyperparameters (initialization variance and learning rate) for large-scale model training, while establishing a robust, theoretically grounded framework for model scaling.

\subsubsection{Model Growth Initialization}

\longcat employs model growth as its initialization strategy, starting from a half-scale model pre-trained on tens of billions of tokens. Among existing model growth methods \citep{chen2015net2net,du2024stacking, wang2023learning, shen2022staged, wang2023lemon, pmlr-v97-gong19a,li-etal-2020-shallow}, we adopt the layer stacking technique \citep{du2024stacking,kim2023solar,li-etal-2020-shallow} to expand parameters and enhance performance. Disregarding the embedding and unembedding processes temporarily, the whole procedure is formulated as:
\begin{align*}
    L_{\text{small}} &= l_1 \circ l_2 \circ \cdots \circ l_n \\
    L_{\text{target}} &= \underbrace{L_{\text{small}} \circ L_{\text{small}} \circ \cdots \circ L_{\text{small}}}_{r}
\end{align*}
where $l_i$ denotes the transformation of the $i$th layer in the model, $r$ denotes the expansion rate, $L_{\text{small}}$ denotes the small model's transformation from token embeddings to final hidden states, and $L_{\text{target}}$ represents the transformation of the target (large) model constructed by stacking $r$ copies of the small model. We use $r=2$ for our architecture.

Through extensive experiments, we consistently observed that models initialized via model growth exhibit a characteristic loss trajectory: an initial increase followed by accelerated convergence, ultimately outperforming randomly initialized baselines. Figure~\ref{fig:model_grrowth_fig2} presents a representative case from our 6B activated model experiments, demonstrating the advantage of model growth initialization.

We conjecture that this improvement arises from two synergistic factors: (1) the faster convergence of smaller models likely provides higher-quality parameter initializations for scaled training, and (2) growth operations potentially serve as implicit regularization against parameter collapse. Experimental evidence further suggests that over-optimizing predecessor models may negatively impact token efficiency in target models, indicating the need for judicious growth timing.

For \longcat initialization, we first train a 14-layer model with identical architecture to the target model, using random initialization on the initial data segment. The trained model is then stacked to create a 28-layer checkpoint, preserving all training states including sample counters and learning rate schedules from the predecessor.

\subsubsection{Training Stability}
We enhance the training stability of \longcat from three perspectives: router stability, activation stability, and optimizer stability.

\paragraph{Router Stability}
A fundamental challenge in training MoE models is router stability, which stems from the tension between two competing gradients: 
\begin{itemize}
    \item The language modeling (LM) loss, driving \textit{expert specialization} (assigning tokens to the most suitable experts),
    \item The auxiliary load balancing (LB) loss, enforcing \textit{routing uniformity} (distributing tokens evenly across experts).
\end{itemize}  
When the LB gradient dominates, router parameters for all experts converge toward similarity, leading to uniform routing decisions regardless of input tokens. This nullifies the benefits of conditional computation and severely degrades model performance.

To diagnose and control this behavior, we propose a monitoring framework with two key metrics:
\begin{itemize}
    \item Router Weight Similarity: Measure the average pairwise cosine similarity between expert weight vectors $\{w_i\}$. A high similarity is a direct indicator that the load balancing loss is excessively dominant.
    \item Gradient Norm Ratio ($R_g$): Quantify the relative influence of the two losses on the batch-averaged expert probability vector $\vec{P}$:
    \begin{equation}
        R_g = \frac{\|\alpha \nabla_{\vec{P}} \mathcal{L}_{\text{LB}}\|_2}{\|\nabla_{\vec{P}} \mathcal{L}_{\text{LM}}\|_2},
    \end{equation}
    where $\mathcal{L}_{\text{LB}}$ is the load balancing loss computed without the coefficient $\alpha$.
\end{itemize}

Guided by this framework, we establish a practical guideline for setting the hyperparameter $\alpha$. The principle is to ensure the load balancing term acts as a regularizer without overwhelming the LM loss. We therefore recommend choosing a coefficient that keeps the $R_g$ below a small threshold (e.g., $R_g < 0.1$).

\begin{figure}[t]
    \centering
    \begin{subfigure}[b]{0.4\linewidth}
        \centering
        \includegraphics[width=\linewidth]{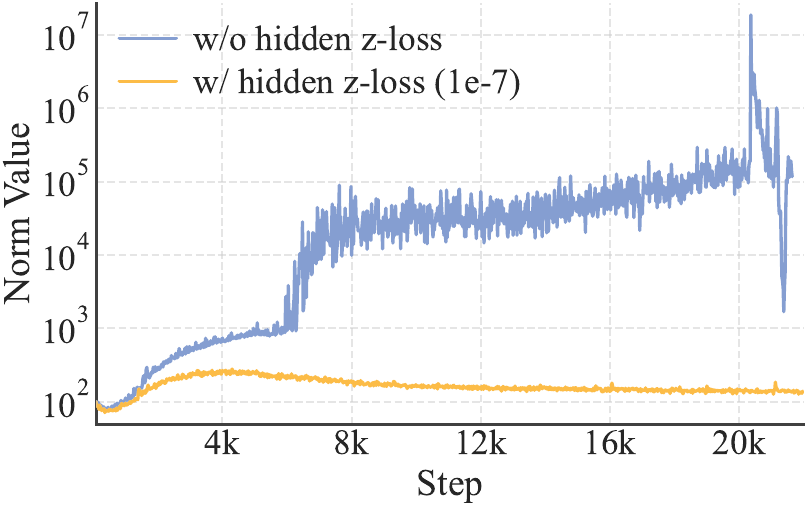}
    \end{subfigure}
    \begin{subfigure}[b]{0.4\linewidth}
        \centering
        \includegraphics[width=\linewidth]{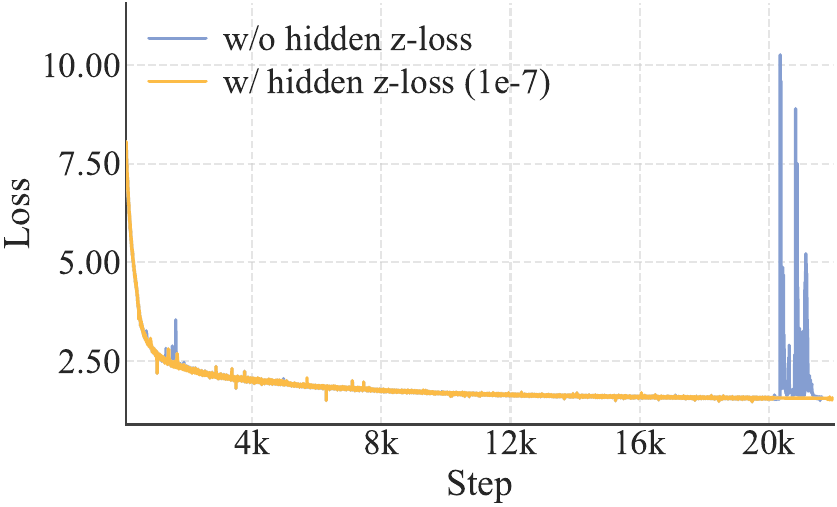}
    \end{subfigure}
    \caption{L2 norm of last layer's hidden states and the training loss for a small model with suboptimal hyper-parameters. The introduction of a negligible-coefficient hidden z-loss stabilizes the norm curve without degrading training loss.}
    \label{fig:norm-loss}
\end{figure}

\paragraph{Activation Stability via Hidden z-loss}
Inspired by the router z-loss \citep{zoph2022stmoedesigningstabletransferable}, we design hidden z-loss to circumvent the widespread occurrence of massive activation \citep{sun2024massiveactivationslargelanguage} during LLM training.  Through empirical observations, we find that such massive activations correlate with severe loss spikes during training, which are associated with optimization instability and potential performance degradation.
Hidden z-loss is mainly used to suppress elements with extremely large magnitudes:
\begin{equation}
    \mathcal{L}_{\text{Z}} = \frac{\lambda}{T}\sum_{t=1}^T(\log \sum_{i=1}^{|z_t|}\exp(\text{abs}(z_t^i)))^2,
\end{equation}
where $\lambda$ is the coefficient to weight this loss,  $z_t$ is the final layer output of the $t$-th token (before final layer norm), $|z_t|$ is the hidden state size, and $\text{abs}(*)$ denotes absolute value function.
As depicted in Figure~\ref{fig:norm-loss}, we found that a very small loss coefficient can significantly suppress the massive activation phenomenon without compromising training loss, thus reducing the risk of numerical errors  during BF16 training.

\paragraph{On the Practical Configuration of Adam's Epsilon}
As model scale increases, the epsilon ($\varepsilon$) parameter in the Adam optimizer, traditionally treated as a minor constant for numerical stability, emerges as a critical hyperparameter. \citet{olmo20242} demonstrated that setting it to 1e-8 yields superior results compared to the default value of 1e-5. This heightened sensitivity primarily stems from two factors: (1) large-scale models typically employ smaller parameter initializations, and (2) they utilize substantially larger batch sizes during training. When using default $\varepsilon$ values, the parameter's magnitude may become comparable to or even exceed the typical scale of gradient second moments, thereby disrupting the optimizer's adaptive mechanism.

\begin{figure}[t]
    \centering
    \includegraphics[width=0.7\linewidth]{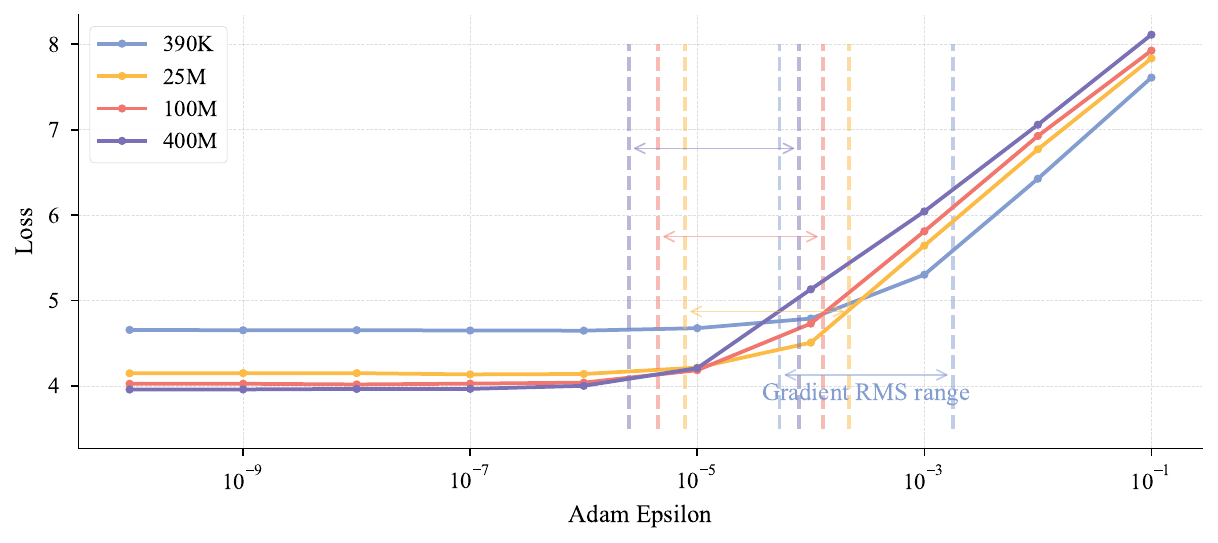}
    \caption{Exploring the impact of the Root Mean Square (RMS) norm of gradients and epsilon on loss across different model sizes. The ``Gradient RMS range'' denotes the span between the maximum and minimum gradient RMS values for different weights in the model. As the model size increases (ranging from 390K to 400M parameters), the gradient RMS becomes smaller. When epsilon approaches the range of the gradient RMS, a rapid deterioration in loss is observed.}
    \label{fig:adam}
\end{figure}

As illustrated in Figure~\ref{fig:adam}, our empirical analysis tracking the Root Mean Square (RMS) norm of gradients reveals two key findings: (1) Threshold effect: Significant performance degradation occurs when $\varepsilon$ approaches the observed gradient RMS norm; (2) Lower bound stability: Once $\varepsilon$ is reduced below this critical threshold, further decreases have a negligible impact on model performance. Consequently, we recommend setting $\varepsilon$ to a small value (several orders of magnitude smaller than the expected gradient RMS norm). In \longcat, we adopt $\varepsilon$=1e-16, a configuration that ensures numerical stability while preserving the optimizer's adaptive properties.

\subsection{General Pre-Training}
We first conduct a general pre-training stage to ensure overall model ability. A multi-phase pipeline is designed to ensure data quality and diversity. The main phases include:

\textbf{Content Extraction} We utilize a customized version of trafilatura~\citep{barbaresi-2021-trafilatura} for general web content and a dedicated process for STEM material to correctly parse complex elements like formulas, code, and tables.

\textbf{Quality Filtering} A two-step filtering approach is applied. An initial classifier removes clearly low-quality documents, followed by finer-grained screening based on metrics like fluency and content completeness.

\textbf{Deduplication} We apply an efficient MinHash implementation for large-scale deduplication, supplemented by a strategy to identify and handle repetitive web templates for more accurate document-level deduplication.

The final data mixture process adopts a two-stage schedule, progressively increasing the proportion of high-quality reasoning data (e.g., STEM and code).

\begin{itemize}
    \item Stage 1: For general-purpose data, we employ an instance-level data mixing strategy that balances data quality and diversity described in SampleMix~\citep{samplemix}, where we compute an initial sampling distribution using quality and diversity scores, and further adjust the tendency of the distribution based on fine-grained domain and writing style labels. Redundant low-value domains (e.g., advertisement, sports, hiring) are downsampled, while reasoning-rich domains (e.g., science) are upsampled.

    \item Stage 2:  We prioritize reasoning-intensive domains in this phase, with STEM and code comprising 70\% of the final mixture. Preliminary experiments showed that abrupt reductions in general-domain data temporarily degrade model capabilities. Thus, we implement gradual code proportion increases, guided by continuous perplexity monitoring on external validation sets to ensure smooth transitions without compromising general performance.
\end{itemize}

\subsection{Reasoning and Coding Enhancement}
To further enhance the model's reasoning and coding capabilities and establish a robust base model with substantial potential for subsequent post-training, we conduct a mid-training stage utilizing high-quality relevant data generated through a combination of pretraining data retrieval and data synthesis.

The systematic synthetic data workflow is introduced to optimize data quality and diversity through three key mechanisms: (1) Knowledge graph traversal and node combination to ensure conceptual complexity and domain coverage; (2) Multi-stage iterative refinement to progressively improve difficulty levels and Chain-of-Thought (CoT) reasoning quality; (3) Dual-modality generation and verification (textual and computational) to guarantee mathematical accuracy and solution validity. 
Careful quality control is conducted combing both rule-based and model-based filters, and the final dataset comprises hundreds of billions of tokens.

\subsection{Long Context Extension}
We implement a two-stage context length extension strategy to meet the requirements for subsequent long-context reasoning and agentic training. In the first stage, the context window expands from 8k to 32k tokens using 80B training tokens, with RoPE's base~\citep{su2024roformer} frequency raised from 1,000,000 to 5,000,000. In the second stage, we further extend it to 128k tokens through an additional 20B tokens, increasing the base frequency to 10,000,000.

The training corpus is built upon naturally occurring long-text data, such as high-quality books and novels. 
Additionally, we developed a systematic approach to organize repository-level source code to improve the model's long-context capabilities. 
We carefully selected high-quality repositories and applied a multi-stage filtering process to remove non-textual content, build artifacts, and auto-generated code, resulting in a curated 20B-token dataset for long-context pre-training.

To ensure that the model's general capabilities remain stable during the length extension, we adopt a data mixture strategy identical to that of our main pre-training phase and augment this mixture with an additional 25\% of long-context data to enhance the model's long-context performance.

\subsection{Decontamination}

We perform rigorous decontamination on all training data to prevent data leakage from test sets of common benchmarks. For web and code data, we remove documents containing any 13-gram overlap with predefined test sets. For synthetic data and question-answering pairs, we employ a stricter strategy based on semantic similarity using BGE-m3 \citep{bge-m3} embeddings. Documents are discarded if they meet either of the following criteria: (1) Semantic similarity score > 0.9 to any test case; (2) Lexical overlap (measured by sparse embeddings) combined with a similarity score between 0.7–0.9.

\subsection{Evaluation}
\input{base_model_evaluation}

\section{Post-Training}

We implement a conventional multi-stage post-training framework to augment the base model's performance across diverse domains, ranging from sophisticated reasoning, coding and agentic tool use tasks to general-purpose capabilities. During this process, we observed that the limited availability of high-quality problem sets is a significant bottleneck across all domains. In the subsequent sections, we present key insights derived from our post-training methodology, organized into three distinct phases: (1) Reasoning and coding, (2) Agentic tool use, and (3) General capability.

\subsection{Reasoning and Coding}
\paragraph{Mathematics}

To generate high-quality and novel problems, we use a persona~\citep{ge2024scaling}, self-instruct~\citep{wang2022self} paradigm. This process is guided by a comprehensive mathematical framework that spans topics from elementary to advanced levels. We leverage a diverse set of math-related ``expert'' personas to ask questions, steering LLMs to synthesize queries that cover underrepresented subjects. Each query is structured to elicit Chain-of-Thought (CoT) reasoning, promoting step-by-step problem-solving in the generated answers. Details of persona curation and answer verification are as follows:

\begin{itemize}
    \item Persona Curation: The personas are constructed from multiple sources: we generate them from our high-quality pretraining data, derive them from existing math queries, and incorporate relevant collections from Persona Hub. Each persona is systematically labeled by its STEM discipline. To ensure maximum diversity and alignment with our subject framework, we use the MinHash algorithm to select the final set of personas for query generation.
    \item Answer Verification: We employ a two-stage process to ensure the accuracy of the synthesized solutions: (1) We generate answers for each problem using several different LLMs and select the most consistent solution as the final answer. (2) We train a generative reward model, specifically enhanced with reasoning data, to automatically score and verify the logical soundness of the problem-solving steps.
\end{itemize}

\paragraph{Coding}

We assemble a diverse set of coding queries from multiple sources, including public datasets, queries generated from GitHub code snippets \citep{SelfCodeAlign_nips} and coding-related forums, as well as queries evolved using the Code Evol-Instruct method \citep{WizardCoder_iclr}. The data distribution is balanced according to topic diversity and difficulty. Specifically, we train a model to select queries that are clear, consistent, and correct, with sufficient explanatory detail, and implement a filtering pipeline to eliminate responses containing garbled content, repetitive patterns, or logical errors. For software engineering tasks, we curate and validate ten thousands of Docker images containing test cases. Each image is used to verify whether model-generated code can resolve specific issues in the corresponding repository. We develop an agent-based system that leverages various tools to autonomously analyze code structures, identify relevant files, fix bugs, and implement new features. This process yields thousands of successful trajectories that pass all test cases, thereby enhancing the model’s ability to autonomously solve real-world software engineering problems.

\paragraph{Logical Reasoning}

We construct logical reasoning datasets covering deductive, hypothetical, and inductive reasoning, which include tasks such as LogicPro~\citep{jiang-etal-2025-logicpro}, PODA~\citep{wang2025PODA}, and Zebra-style logic puzzles. To manage difficulty, we first use the Pass@k metric for an initial balance, then filter out intractable problems where advanced thinking models failed. We also convert multiple-choice questions to a fill-in-the-blank format to mitigate random guessing. The evaluation of responses focused on four key areas: (1) correctness of the final answer; (2) completeness and clarity of reasoning; (3) avoidance of excessive repetition; and (4) consistent use of language.

\subsection{Agentic Tool Use}
We define agentic tasks as complex problem resolution through systematic environment interaction. In this paradigm, models must iteratively analyze existing information and determine when environmental interaction is needed. Specifically, within the agentic tool utilization framework, the environment comprises user and tools with distinct characteristics. User functions as an autonomous information-providing entity without upstream or downstream dependencies, but exhibit reluctance to be disturbed and non-spontaneous information disclosure. Consequently, models must minimize user queries while employing strategic questioning techniques to elicit maximally precise information when interaction becomes necessary. Tools can be invoked extensively with high frequency, but exhibit intricate interdependencies. From this perspective, excluding domain-specific expertise such as advanced programming capabilities or mathematical computation, we attribute task difficulty escalation to three factors:
\begin{itemize}%
\item \textbf{Information processing complexity} 
Models must engage in sophisticated reasoning processes to integrate and transform information into required components.
\item \textbf{Tool set complexity} 
By modeling the tool set as a directed graph based on intertool dependencies, complexity can be quantitatively characterized by the graph's node cardinality and edge density.
\item \textbf{User interaction complexity}
Models must learn to engage in multi-round strategic questioning with minimal frequency, adapting to various conversational styles, levels of communication willingness and patterns of information disclosure, thus facilitating effective user interaction while ensuring adequate information acquisition.
\end{itemize}

Building on these insights, we construct a multi-agent data synthesis framework that generates high-quality challenging tasks by systematically addressing three complexity dimensions critical for agent training: (1) tool set complexity, (2) information processing complexity, and (3) user interaction complexity. The framework comprises the following specialized agents:

\begin{itemize}
\item \textbf{UserProfileAgent} Beyond generating fundamental user profiles encompassing personal information and preferences, we further implement controls over user conversational styles, communication willingness levels, and information disclosure patterns to more accurately simulate authentic user interaction scenarios while simultaneously enhancing task complexity.

\item \textbf{ToolSetAgent} To maximize data diversity and prevent overfitting to specific scenarios, we adopt an approach analogous to Kimi-K2~\citep{team2025kimi}, enumerating 40 distinct domains and subsequently leveraging models to enumerate 1,600 applications. Based on these applications, we construct 80,000 mock tools, forming an extensive tool graph. Through random walk methodologies, we systematically sample subgraphs with predetermined node quantities from this comprehensive tool graph, and hence tool graph complexity is controlled via node quantity.

\item \textbf{InstructionAgent} The difficulty of reasoning is quantified in the following dimensions: constraint complexity, quantity of reasoning points, and length of the reasoning chain. The model is required to generate instructions that comprehensively describe complete tasks based on the tool set extracted by the \textbf{ToolSetAgent}.

\item \textbf{EnvironmentAgent} We augment environmental information including item details, location specifics, temporal parameters, and meteorological conditions based on content generated by the \textbf{UserProfileAgent} and \textbf{InstructionAgent}. Additionally, we introduce confounding elements for items and locations to further increase reasoning complexity.

\item \textbf{RubricAgent} We construct a comprehensive series of specific checklists based on various task-related information. During final evaluation, considering the long-context characteristics inherent to agentic tasks, we employ a sliding window approach to assess the entire trajectory, continuously updating the completion status of checklist items.

\item \textbf{ValidatorAgent} and \textbf{DeduplicatorAgent} We check the quality of our final tasks from several angles and remove any that are too similar. This process ensures we have a diverse and high-quality set of tasks.
\end{itemize}

With these high-quality challenging tasks, we further conduct rigorous response selection to construct our cold start training set with an appropriate quantity, revealing diverse patterns and preserving high exploration ability. We also carefully select a subset of these generated task for further post-training procedure, to make sure each task worth massive exploration.

\subsection{General Capability}

\paragraph{Instruction-following} 

We curate both single-turn and multi-turn instruction-Following datasets, with varying levels of constraint complexity and quantity. For multiple-constraint queries, we adopt the insight from \citet{ye2025multi} to filter queries with low semantic quality or constraint conflicts. For different query types, we employ verifiable rules, model-based verification, and customized strategies to ensure responses satisfy all constraints. Additionally, we compile critique datasets targeting challenging tasks to enhance the model's critical thinking abilities \citep{Critique_ft}. We observe that certain constraint types are inherently difficult to follow, making direct generation of valid query-answer pairs unreliable. To address this, we propose a reverse prompt generation strategy: generating queries from predefined answers guaranteed to meet constraints.

\paragraph{Long Context}
To enable the model to identify and analyze relevant information within complex, lengthy contexts, we develop three types of long-sequence datasets: reading comprehension, table-based question answering, and custom-designed tasks. 
To facilitate the learning of salient information in long sequences, we aggregate topically related context segments for data construction.
We specifically enhance the model’s multi-hop reasoning, multi-turn dialogue, and complex calculation abilities. To mitigate hallucination when confronted with an incomplete context, we optimize the model’s refusal capabilities, thereby improving its awareness of knowledge boundaries and limitations.

\paragraph{Safety}
Building on the framework of \citet{rule_safety_nips} and aligned with our internal content guidelines, we develop a content safety policy that categorizes queries into more than 40 distinct safety categories across five response types: \textit{comply, comply with guideline, soft refuse, soft refuse with guideline, or hard refuse}. Explicit criteria ensure consistent, safety standards-compliant responses for each response type. This system operates as a context-aware data synthesizer through two stages: (1) Query Classification: Queries from diverse sources (open-domain corpora, internal business risk reports, government Q\&A, and adversarial LLM-synthesized red-team content) are classified into safety categories using automated labeling with human verification. (2) Response Mapping \& Optimization: Classified queries are mapped to response types and generate optimized, type-specific responses that undergo human evaluation before serving as training targets.

\subsection{Evaluation}
\input{chat_model_evaluation}

\section{Training Infrastructures}
The core design principle of our training infrastructure is scalability with precision. We developed a systematic method to verify operator precision and embedded online Silent Data Corruption (SDC) detection into idle computation phases to minimize numerical errors. To guarantee reproducibility and ensure consistent results between small-scale experiments and full-scale training, we enforced determinism across all computation and communication operators. This enabled bitwise-aligned loss values across multiple re-runs of any training step.

With correctness ensured, we focused on accelerating training efficiency. Wall-clock time is critical for rapid algorithm iteration, yet single accelerator provides limited capability. We therefore scaled training across tens of thousands of accelerators, confronting challenges in scalability and stability. Through model–system co-design, multi-dimensional parallelism, and fully automated fault detection and recovery, we achieved near-linear scaling and 98.48\% availability, completing training within 30 days.

\subsection{Numerical Precision Control and Fault Detection}
\textbf{ULP Evaluation} Floating-point errors are influenced by multiple factors, even varying between \chip of the same vendor across generations. To quantify and mitigate these errors, we adopt ULP (Unit in the Last Place) as a metric, where ULP error measures the deviation of accelerator BF16 results from CPU FP32 ground truth. A zero ULP error indicates perfect accuracy, while larger values imply worse precision. We collect all operator types and shapes used in training and compare their ULP errors. Table~\ref{tab:gemm ulp} shows the ULP error for GEMM between two solutions.

\begin{table}[h!]
    \centering
    \caption{GEMM Precision Comparison (ULP)}
    \begin{tabular}{ccccccc}\toprule
 & & & \multicolumn{2}{c}{Solution 1}& \multicolumn{2}{c}{Solution 2}\\\midrule
         Case&  Output Shape&  Value Range&  Max&  Min&  Max&  Min\\
         1&  [1024,1536]&  [-5,5]&    2292&  -568&  112&  -100\\
         2&  [1024,576]&  [-5,5]& 65362&  -82046& 6.5&  -9  \\
         3&  [1024,16384]&  [-19,15]&   544&  -104& 224&  -112 \\
         4&  [1024,12288]&  [-4,4]& 202&  -88& 72&  -41 \\
         5&  [1024,6144]&  [-1,1]&  5376&  -1376& 304&  -224\\
         6&  [1024,24576]&  [-5,5]&  7200&  -510& 104&  -294\\
 7& [1024,131072]& [5,5]& 8128& -6976& 2528& -368 \\
 8& [1024,6144]& [-1,1]& 5344& -8064& 80& -258 \\ \bottomrule
    \end{tabular}
    \label{tab:gemm ulp}
\end{table}

\textbf{SDC Detection Mechanism} SDC faults are typically unavoidable in large-scale training and can severely degrade model performance by altering data without system warnings. To address this, we implement an efficient on-chip in-place operator recomputation mechanism. Specifically, we find that the backward computation for FlashAttention Gradients (FAG) is most sensitive to SDC because it simultaneously mixes tensor and vector computations. Bitwise differences between recomputed results indicate potential SDC risks. The detection computations are orchestrated within compute streams, and the recomputation interval is manually adjustable, enabling a flexible trade-off between detection coverage and computational cost.

Notably, operator precision control is necessary but insufficient for ensuring model accuracy. Experiments with different operator implementations may show training loss discrepancies within 1e-3$\sim$1e-4 yet exhibit larger than 5 pp variation on benchmarks. Cost-effectively evaluating the impact of operator precision errors on model performance remains an open challenge.

\subsection{Kernel Optimization for Determinism and Performance}
Determinism serves as the gold standard for computational correctness, eliminating floating-point errors as experimental variables. However, achieving determinism often incurs significant performance overhead. We address this through kernel redesigns, maintaining deterministic computation and communication throughout \longcat's training.

\textbf{Deterministic FAG} The default FAG implementation is non-deterministic because $dQ$, $dK$, and $dV$ are reduced along different dimensions, where atomic addition lacks order preservation. We develop an efficient deterministic FAG kernel using limited extra workspace to accumulate tiles in a deterministic order. With co-optimizations including double-buffer pipelining, tuned tiling schedules, and load balancing, our implementation achieves 1.6x the performance of the original deterministic version and 0.95x that of the non-deterministic version, striking a balance between determinism and efficiency.

\textbf{Deterministic ScatterAdd} ScatterAdd in backward passes is essential for gradient aggregation but suffers from input-output operand counts mismatches. The default implementation enforces sequential execution within a single compute unit, causing up to 50x slowdown. We propose a hierarchical reduction algorithm that parallelizes gradient aggregation across all available processors, achieving performance parity with the non-deterministic version.

\textbf{Optimized Grouped GEMM} Grouped GEMM's performance is critical given its high computational volume but low compute density versus dense GEMM. We optimize it via: (1) Double-buffer pipelining to overlap computation, memory I/O, and epilogue; (2) Diagonal tiling to mitigate L2 cache conflicts; (3) HBM bandwidth control via compute unit limits to overlap Grouped GEMM with dispatch/combine communication. These optimizations yield 5\%–45\% speedups over the default version.

\textbf{Fused GemmAdd} The $dw$ computation suffers bandwidth-bound bottlenecks during gradient accumulation. We fuse FP32 addition into the GEMM epilogue, avoiding intermediate write-backs and hiding addition within tile GEMM pipelines. This significantly reduces latency and eliminates the precision loss caused by the conversion of BF16 data to HBM, achieving a speedup of 3.12x to 3.86x on the fused GroupedGemmAdd benchmark.

Furthermore, we re-implement IO-bound kernels (e.g., MoE layer permute/unpermute) with integrated functionalities like drop-token and zero-computation experts handling, ensuring both determinism and performance.

\subsection{Distributed Strategy for Large-scale Training}
The training architecture is centered on Expert Parallelism Groups (EP), each comprising 32 \chip. Within an EP Group, the attention layer employs Context Parallelism (CP=8) instead of Tensor Parallelism (TP) to minimize communication overhead, and the FFN layer uses EP partitioning without TP.
Multiple EP groups are scaled across Pipeline Parallelism (PP) and Data Parallelism (DP) dimensions.

\begin{figure}[h]
\begin{center}
    \centering
    \includegraphics[width=0.99\linewidth,]{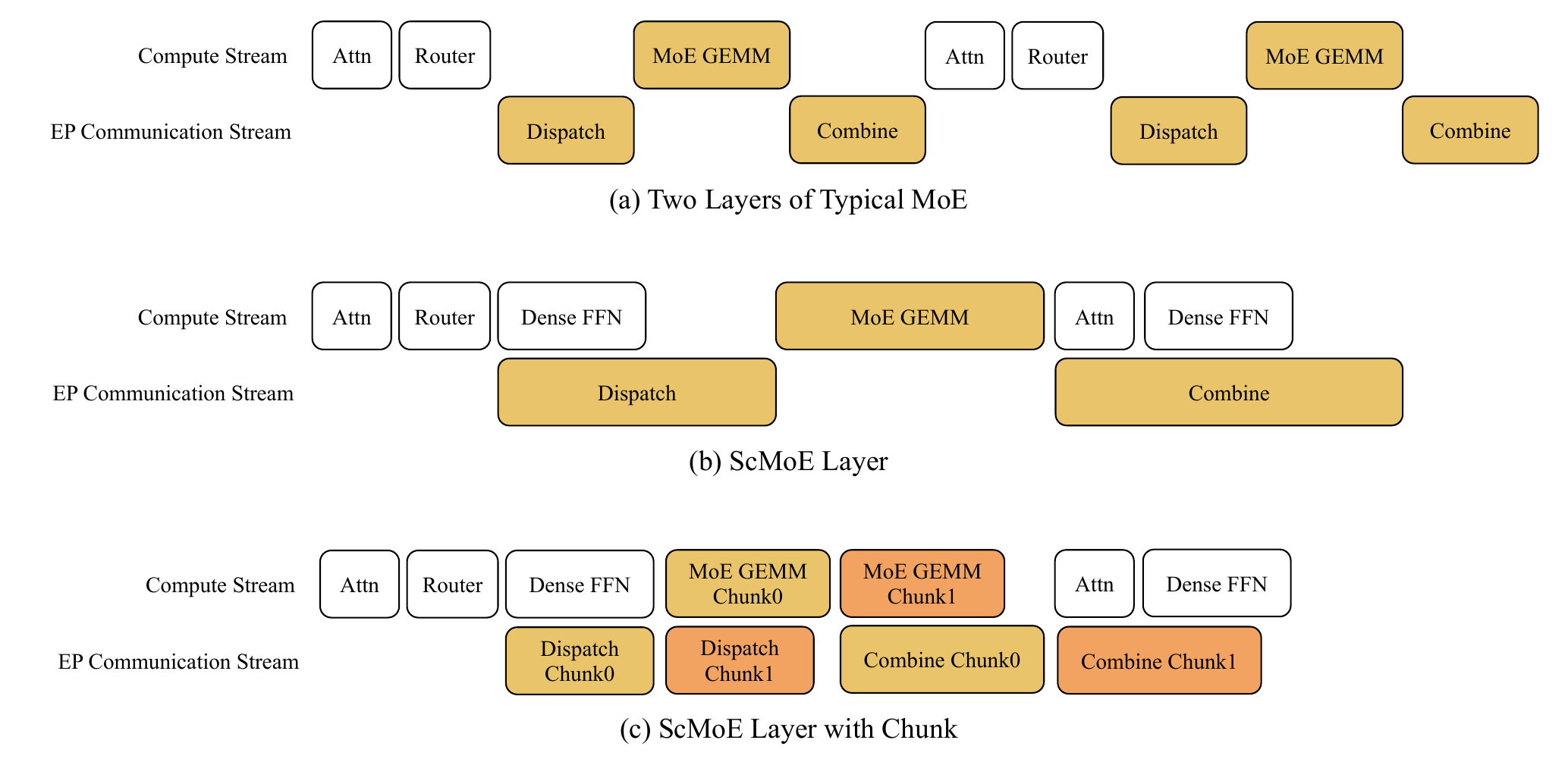}
    \caption{These architectures have the same total and activated number of experts. ScMoE with chunk achieves the highest efficiency because more communication is overlapped by computation.}
    \label{fig:ScMoE-arch}
\end{center}
\end{figure}

Expert parallelism (EP) is adopted to reduce static memory usage, including weights and optimizer states. However, EP inherently introduces costly dispatch and combine communication operations. To mitigate this, \longcat adopts the ScMoE structure, which enables dispatch/combine communication to overlap by more computation in a single batch. Furthermore, the MoE layer is divided into two chunks along the token dimension. These subchunks achieve two objectives: (1) Overlap with the dense FFN computation. (2) Overlap with each other (see Figure~\ref{fig:ScMoE-arch}).

There are two optimized strategies for dispatch/combine communication: (1) All-gather/reduce-scatter kernel with pipeline in the intranode and the internode; (2) Optimized all-to-all kernel. The native all-to-all expands the local data size by top-k times, increasing traffic through the 200Gb/s per accelerator RDMA network. Additionally, all-to-all performance is unstable due to inadequate congestion control. We select pipelined all-gather/reduce-scatter with deterministic as the primary solution, the proportion of time to non-overlapping dispatch/combine communication decreases from 25.3\% to 8.4\% with ScMoE architecture.

Existing pipeline strategies (e.g., 1F1B, interleaved-1F1B, Zero-bubble \citep{qi2023zero}) suffer from imbalanced memory usage across pipeline stages. To this end, we adopt the V-ZB algorithm \citep{qi2024zbv}, which balances memory usage at all stages and reduces peak memory to less than 60GB in the training of \longcat. Additionally, we enable the post-validation strategy from zero bubble, achieving zero theoretical bubbles. A key refinement is replacing inverse operations with backup data from the previous step during optimizer state rollback, preserving numerical bitwise alignment.

\subsection{Reliability and Observability}
Reliability is measured by the proportion of time contributing to the final training trajectory (Availability), where unavailable time includes fault recovery and wasted time between the last checkpoint and fault occurrence. Asynchronous checkpointing reduces training stall to 2$\sim$4 seconds, allowing higher frequency and minimizing fault-induced loss. Combined with online critical log filtering, optimized initialization, and full automation, recovery time is reduced to <10 minutes. These mechanisms achieve 98.48\% availability, with all 20 faults handled automatically without manual intervention.

Observability combines fine- and coarse-grained profiling with a metric platform. Fine-grained PyTorch profiler timelines enable distributed, parallel-aware co-analysis to identify pipeline parallelism "bubbles" and inter-rank communication waits. Coarse-grained monitoring adds low-overhead runtime analysis of stragglers. The metric platform tracks loss, weights, gradients, and activations for rapid model state assessment.

\section{Inference and Deployment}

\longcat employs a model-system co-design, which significantly contributes to its high throughput and low latency. This section focuses on inference optimizations implemented in one of our deployment clusters, presenting methods to simultaneously boost system throughput and significantly reduce latency to 100 TPS on H800. We first present our parallel inference architecture co-designed with the model architecture. Following the inference architecture, optimization methods such as quantization and custom kernel are described. Finally, we present our deployment strategy and performance results.

\subsection{Model-Specific Inference Optimization}

To achieve an efficient inference system, two key challenges must be addressed: (1) Computation and communication orchestration, and (2) KV cache I/O and storage. For the first challenge, existing approaches typically exploit parallelism at three conventional granularities: operator-level overlap like NanoFlow \citep{zhu2025nanoflowoptimallargelanguage}, expert-level overlap represented by EPS-MoE \citep{qian2025epsmoeexpertpipelinescheduler}, and layer-level overlap demonstrated in DeepSeek-V3 TBO~(Two Batch Overlap) \citep{SGLang2025}. \longcat's ScMoE architecture introduces a fourth dimension—module-level overlap—for which we designed the SBO~(Single Batch Overlap) scheduling strategy to optimize both latency and throughput. For the second challenge—KV cache I/O and storage—\longcat\ addresses these issues through architectural innovations in its attention mechanism and MTP structure to reduce the effective I/O overhead.

\subsubsection{Computation and Communication Orchestration}

\begin{figure}[t]
    \centering
    \includegraphics[width=0.9\linewidth]{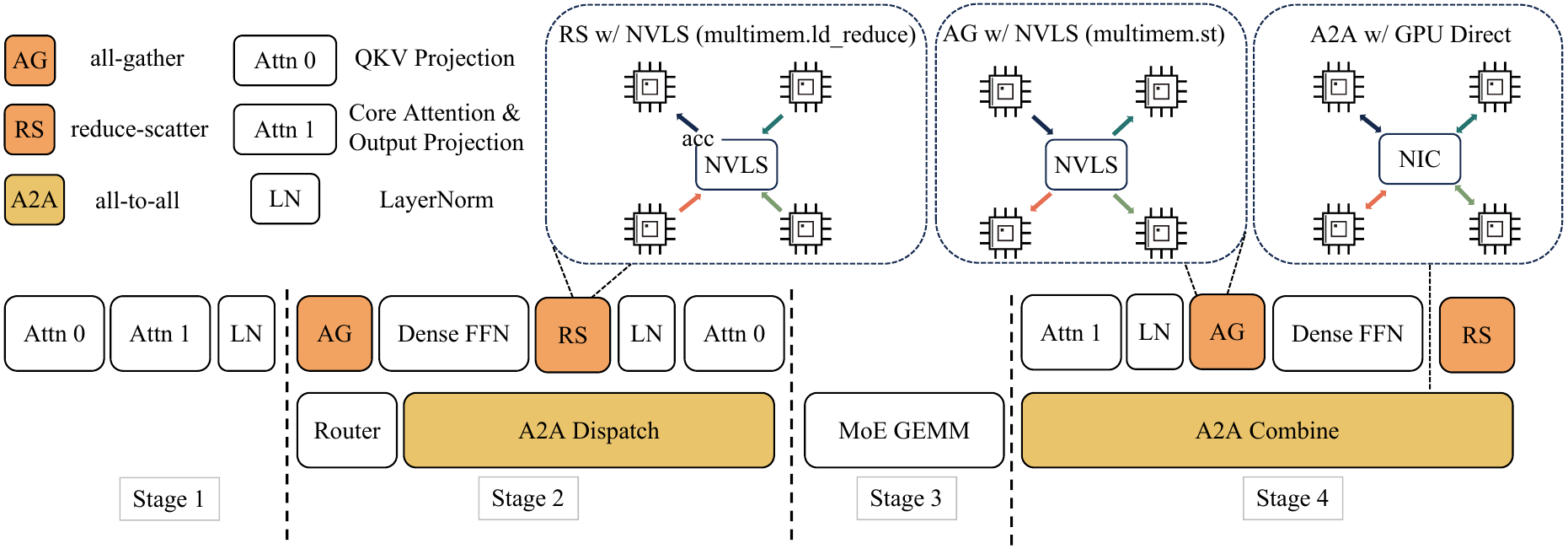}
    \caption{An overview of overlapping strategy.}
    \label{fig:infer-stage-overlap}
\end{figure}

\longcat naturally exhibits computation-communication overlap properties in its structure, which is the key to achieving lower latency while maintaining generation throughput. We carefully design Single Batch Overlap (SBO), a four-stage pipeline execution that uses module-level overlap to fully unleash \longcat's potential as shown in Figure~\ref{fig:infer-stage-overlap}. SBO differs from TBO by hiding communication overhead within a single batch. In SBO, stage 1 requires separate execution because the MLA output serves as input for subsequent stages. In stage 2, we overlap all-to-all dispatch with Dense FFN and Attn 0 (QKV Projection). This overlap is crucial because communication overhead is excessive, prompting us to split the attention process. Stage 3 independently executes MoE GEMM. The latency of this stage will benefit from the wide EP deployment strategy. In stage 4, we overlap Attn 1 (Core Attention and Output Projection) and Dense FFN with the all-to-all combine. This orchestration effectively mitigates the communication overhead, ensuring efficient inference for \longcat.

Additionally, the ScMoE architecture, under the wide EP deployment scheme, facilitates the overlap of intra-node NVLink bandwidth utilization and inter-node RDMA communication through GPUDirect RDMA~\citep{choquette2022nvidia}, thereby improving overall bandwidth efficiency. Dense FFN in ScMoE has a relatively large intermediate size, so TP deployment is employed to minimize memory footprint, necessitating all-gather and reduce-scatter communication before and after Dense FFN, respectively. To reduce this communication overhead, we develop custom kernels and adopt TP2 or TP4 instead of TP8. 

\subsubsection{Speculative Decoding}

\longcat employs MTP as the draft model for speculative decoding. Our optimization framework originates from a systematic breakdown of Speculative Decoding's speedup formulation, as \citet{sadhukhan2025magicdecbreakinglatencythroughputtradeoff} has mentioned:

\begin{equation}
\frac{T^{SD}_{Avg}}{T_T} = \frac{1}{\Omega(\gamma, \alpha)} \left(\frac{\gamma \cdot T_D}{T_T} + \frac{T_V(\gamma)}{T_T}\right), \notag
\label{eq:spec_formula}
\end{equation}

where $T^{SD}_{Avg}$, $T_T$, $T_D$ are expected latency per token for speculative decoding, target model and draft model. $\gamma$ represents number of draft token in one decoding step. $\Omega(\gamma, \alpha)$ is expected accept length for a given step $\gamma$ and acceptance rate $\alpha$. And $T_V(\gamma)$ is expected latency for target verification. Our approach targets three key factors:

\begin{itemize}
\item Expected accept length $\Omega(\gamma, \alpha)$, which is positively correlated with the acceptance rate $\alpha$ of draft tokens. To maximize acceptance rate $\alpha$, we employ MTP. Integrate a single MTP head during late-phase pre-training, achieving approximately 90\% acceptance rate on test sets.
\item  Draft to target cost ratio $\gamma\frac{T_D}{T_T}$, which is dominated by the structures of both target model and draft model. As noted by \citet{liu2024speculativedecodingearlyexitingfaster}, balancing draft quality and speed is critical. To minimize generation overhead while maintaining comparable acceptance rates, \longcat adopts a lightweight MTP architecture with reduced parameters. Our experiments (Table~\ref{table:mtp_speedup}) show that a single dense layer for MTP heads optimizes this trade-off, outperforming ScMoE layers in latency.
\item Target verification to decoding cost ratio $\frac{T_V(\gamma)}{T_T}$. In order to reduce this ratio, we adopt the C2T \citep{huo2025c2tclassifierbasedtreeconstruction} method, using a classification model to filter out tokens that are unlikely to be accepted before verification.
\end{itemize}

\begin{table}[ht]
\caption{Draft token acceptance rate on MT-Bench of different MTP head structures with a 6B activated model. The ratio of MTP head parameters to main model parameters is also reported.}
\centering
\begin{tabular}{cccc}
\toprule
MTP layer & Activated parameters ratio &  Acceptance rate $\alpha$ \\ 
\midrule
Dense layer  & 1.41\%  & 92.1\%   \\ 
ScMoE layer  &  4.17\%  &  92.9\%  \\ 
\bottomrule
\end{tabular}
\label{table:mtp_speedup}
\end{table}

\subsubsection{Reducing KV Cache}

To balance performance and efficiency, \longcat adopts MLA with 64 heads for its attention mechanism, which reduces the computational load of the attention component while achieves exceptional KV cache compression and thus reduces storage and bandwidth pressure. This is crucial for orchestrating \longcat's pipeline, as noted in Figure~\ref{fig:infer-stage-overlap} the model always features an attention computation that cannot be overlapped with communication. Specifically, the MQA-like structure of the MLA absorb method shares KV across the m-dimension (64 heads), aligning with the shape of the WGMMA instruction for maximal hardware utilization. %

\subsection{System-Wide Inference Techniques}

\subsubsection{Minimize Schedule Overhead}

\begin{figure}[ht]
    \centering
    \includegraphics[width=0.9\linewidth]{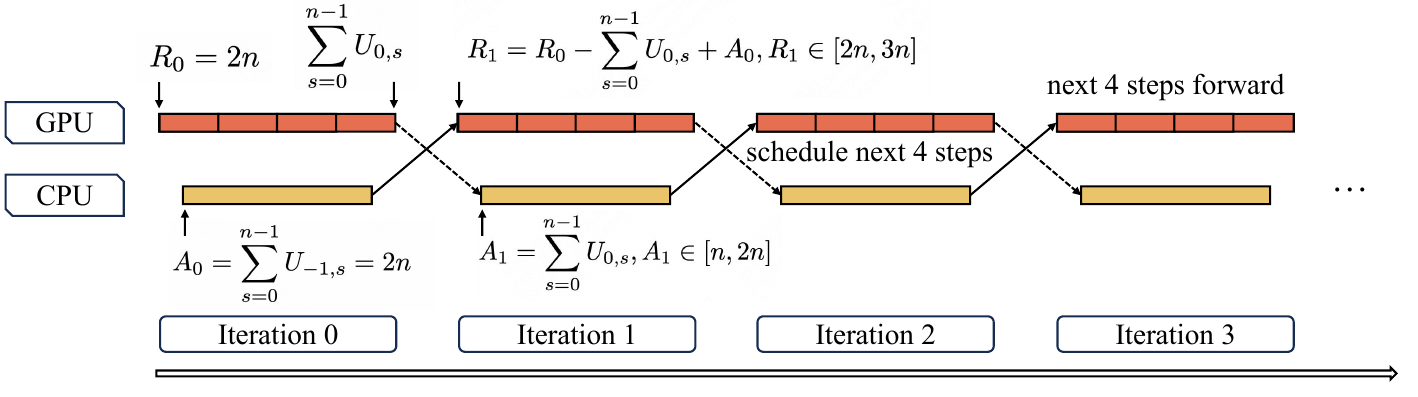}
    \caption{Multi-step overlapped scheduler (4 steps as a example here).}
    \label{fig:infer-overlap}
\end{figure}

The decoding phase in LLM inference systems can become launch-bound due to kernel launch overhead. This issue is exacerbated when introducing speculative decoding---particularly with \longcat's lightweight MTP, where separate scheduling of verification kernels and draft forward passes introduces significant overhead. To mitigate this, a TVD fusing strategy is used to fuse Target forward, Verification, and Draft forward into a single CUDA graph.
To further improve GPU utilization, we implement an overlapped scheduler. However, experimental results reveal that the low latency of \longcat's forward pass renders a single-step pre-schedule strategy insufficient to fully eliminate scheduling overhead. As shown in Figure~\ref{fig:infer-overlap}, we introduce a multi-step overlapped scheduler to launch the kernel for multiple forward steps in a single schedule iteration. This approach effectively hides CPU scheduling and synchronization within the GPU forward process, ensuring continuous GPU occupancy. 

In a multi-step overlapped scheduler, we need to dynamically pre-allocate KV cache slots for multiple future steps  without prior knowledge of the accept length of speculative decoding in previous iterations. An important issue is whether multi-step overlapped scheduling causes divergent KV cache allocation.
We illustrate this with $MTP = 1$ and the number of steps, $n = 4$. 
Let $R_{i}$ represents available KV entries during the GPU's $i$-th iteration forward pass, thus $R_{0}=(MTP + 1) \times n=2n$. $U_{i, s} \in [1, 2]$ represents the accept length in the $i$-th iteration for the $s$ step, with the initial value $U_{-1, s} = 2$.
Then, while the GPU is performing the $i$-th iteration of forward computation, the scheduler pre-allocates the KV cache slots needed for the $(i+1)$-th forward iteration based on the accept length in the $(i-1)$-th forward iteration, where $A_{i}$ represents the KV cache slots that is allocated. Formally:

\begin{align*}
A_{i} &= \sum_{s=0} ^{n-1} {U}_{i-1, s}, \ i \geq 0\\
R_{i} &= R_{i-1} - \sum_{s=0} ^{n-1} {U}_{i-1, s} + A_{i-1}, \ i \geq 1
\end{align*}

By induction, we obtain the closed-form expression:
\begin{align*}
    R_{i} &= 4n - \sum_{s=0} ^{n-1} {U}_{i-1, s}, \ i \geq 1
\end{align*}

which means:
\begin{align*}
    R_{i} &\in [2n, 3n], \ i \geq 1
\end{align*}

Through mathematical induction, this ensures safe KV cache allocation for the next iteration even without knowing the current iteration's accept length, while guaranteeing convergence in allocated KV cache size.

\subsubsection{Custom Kernel}

The autoregressive nature of LLM inference creates distinct efficiency challenges. The prefilling phase is compute-bound, and methods like chunk-prefill \citep{agrawal2023sarathi} regularize data for optimal processing. In contrast, the decoding phase is often memory-bound due to small, irregular batch sizes from traffic patterns, which hurts kernel performance. Therefore, optimizing these specific cases is crucial for minimizing Time-Per-Output-Token (TPOT).

\textbf{MoE GEMM} Existing libraries like DeepGEMM \citep{zhao2025deepgemm} map model weights to right-hand matrices (B in A×B=C) aligned with k/n dimensions, while input activations become left-hand matrices mapped to m/k dimensions, where m represents token count. This conventional approach requires padding when token counts fall below m's 64-element minimum. 
To address this inefficiency, we leverage the SwapAB \citep{dege2025flashmlaetapefficienttransposeattention} technique: treating weights as left-hand matrices and activations as right-hand matrices. By exploiting the n-dimension's flexible 8-element granularity, SwapAB maximizes tensor core utilization.

\textbf{Communication Kernels} The inference system leverages NVLink Sharp's hardware-accelerated broadcast (multimem.st) and in-switch reduction (multimem.ld\_reduce) to minimize data movement and SM occupancy, as shown in Figure \ref{fig:infer-stage-overlap}. 
By using inline PTX assembly, the reduce-scatter and all-gather kernels enable high-efficiency data transmission. These kernels support both uniform and nonuniform token distributions across GPUs, and consistently outperform NCCL \citep{nccl-2.21.5} and MSCCL++ \citep{shah2025mscclrethinkinggpucommunication} across 4KB to 96MB message sizes, using only 4 thread blocks.

\subsubsection{Quantization}

\longcat employs the same quantization scheme as DeepSeek-V3, using fine-grained block-wise quantization: activations per [1,128] blocks and weights per [128,128] blocks. Besides, to achieve an optimal performance-accuracy trade-off, we applied layer-wise mixed-precision quantization based on two methodologies: The first scheme follows our approaches in FPTQ \citep{li2023fptqfinegrainedposttrainingquantization} and Super-Expert \citep{su2025unveilingsuperexpertsmixtureofexperts}, where we observed that certain linear layers (particularly Downproj) exhibited input activations with extreme magnitudes reaching $10^6$. The second scheme involves computing block-wise FP8 quantization errors (both relative and absolute) layer by layer, which revealed significant quantization errors in specific expert layers. By taking the intersection of both schemes, we achieved substantial accuracy improvements.

\subsection{Deployment and Performance}

\subsubsection{Measured Performance}

\begin{table}[h]
\centering
\caption{Performance of \longcat under different settings.}

\begin{tabular}{llllrr}
\toprule
\textbf{Model} & \textbf{Attention} & \textbf{Avg Context} & \textbf{\#Hopper GPUs} & \textbf{TGS} &\textbf{TPS/u} \\
\midrule
DeepSeek-V3-profile & bf16  & 4096 & 128 & 2324 & 20 \\
DeepSeek-V3-blog & bf16  & 4989 & 144 & 1850 & 20 \textasciitilde 22 \\
\longcat & bf16  & 5000 & 128 & 3785 & 35 \\
\longcat & bf16  & 5000 & 128 & \textbf{2205} & \textbf{68.9} \\
\longcat & bf16  & 5000 & 128 & \textbf{804} & \textbf{100.5} \\
\longcat & fp8  & 5000 & 128 & 4230 & 26.4 \\
\longcat & fp8  & 8192 & 128 & 3240 & 33.8 \\
\bottomrule
\end{tabular}

\label{table:precision}
\end{table}

To enable independent optimization of prefilling and decoding phases, PD-Disaggregated architecture is adopted. A key challenge in this design is the overhead of transmitting KV caches from prefilling to decoding nodes. To mitigate this, we implement layer-wise transmission, which significantly reduces Time-To-First-Token (TTFT) under high QPS workloads. For prefilling and decoding nodes, the minimum deployment unit consists of 2 nodes with 16 H800-80GB GPUs. Meanwhile, wide EP is deployed with DeepEP~\citep{deepep2025} to minimize communication overhead. Besides, we modify DeepEP and EPLB~(Expert Parallelism Load Balancer) to support zero-computation experts, where the outputs of zero-computation experts can be obtained without communication. 

Table~\ref{table:precision} compares the throughput and latency of \longcat with DeepSeek-V3 (DeepSeek-V3-profile from \citet{DeepSeekprofile}, DeepSeek-V3-blog from \citet{DeepSeekblog} ), where TGS (token per GPU per second) represents generation throughput per device (higher values indicate lower cost), and TPS/u (tokens per second per user) represents the generation speed for one user (higher values are better). During testing, the steady-state generation throughput under a given sequence length is used for calculation. \longcat achieves higher generation throughput and faster generation speed across different sequence lengths.

In Agent applications based on the ReACT \citep{yao2023reactsynergizingreasoningacting} pattern, completing a single task requires multiple rounds of model interactions, where interaction latency directly impacts user experience. Analysis of typical Agent invocation patterns reveals differentiated speed requirements for model outputs:
\begin{itemize}
    \item Reasoning content (user-visible): consisting of cognitive processes and explanations, must match human reading speed (~20 tokens/s).
    \item Action commands (user-invisible): structured data such as function names and parameters, typically 30\textasciitilde 100 tokens, yet directly affect tool invocation startup time—demanding the highest possible speed.
\end{itemize}

To address this scenario, \longcat achieves a generation speed of nearly 100 tokens/s for action commands. Under a cost assumption of \$2 per hour for an H800 GPU, this translates to a price of \$0.7 per million output tokens. This performance constrains the single-round tool-call latency to under one second, thereby significantly enhancing the interactivity of Agent applications.

\subsubsection{Theoretical Performance}

Figure~\ref{fig:infer-stage-overlap} shows that \longcat's latency is primarily determined by three components:
\begin{itemize}
\item MLA: Its time consumption cannot be reduced by increasing the number of EP.
\item All-to-all dispatch/combine: Both are constrained by single-device batch size and topk.
\item MoE: Its time consumption in the memory-bound region decreases with increasing EP count.
\end{itemize}

Assuming the number of EP is 128, MLA uses DP for DeepSeek-V3 and \longcat, GQA uses TP4 for Qwen3-235B-A22B as it has 4 kv heads, and the batch size per device is 96. Actually, the GQA feature of Qwen-235B-A22B results in a relatively high memory footprint for its KV cache, making it difficult to achieve a per-GPU batch size of 96 in practice. The assumption that it can reach this value is made here solely for the purpose of theoretical analysis. As pointed out by~\citep{flashmla2025}, FlashMLA can achieve up to 660 TFlops on NVIDIA H800 SXM5 GPUs; ~\citet{deepep2025} indicates that DeepEP bandwidth can reach 40GB/s. Both of these metrics were utilized in our computations. Assuming the cost for per H800 is \$2 per hour. Considering MTP=1 with an acceptance rate of 80\%, we can calculate the theoretical time consumption and cost of each module in one layer of DeepSeek-V3, Qwen3-235B-A22B and \longcat, as listed in  Table~\ref{table:per_layer_time}. For Qwen3-235B-A22B, which does not natively support MTP, we assume a speculative sampling strategy with a comparable acceptance rate.

\begin{table}[h]
    \caption{Theoretical decoding time and cost of different models.}
    \centering
    
    \begin{tabular}{cccc}
        \toprule
          & DeepSeek-V3 & Qwen3-235B-A22B & \longcat \\
         \midrule
         MTP & w/ & w/o & w/ \\
         n\_layer            & 61  & 94  & 28 \\
         batch per device & 96 & 96 & 96 \\
         \midrule
         \multicolumn{4}{r}{Time cost of different modules in one layer} \\
         \midrule
         attention          & 471 us & 314 us & 264 us \\
         all-to-all dispatch & 275 us & 157 us  & 236 us \\
         MoE                 & 77 us  & 29 us  & 60 us \\
         all-to-all combine  & 551 us & 315 us & 472 us \\
         \midrule
         \multicolumn{4}{r}{TPOT and Price}\\
         \midrule
         overlap strategy    & TBO & TBO & SBO \\
         TPOT (ms)           & 30  & 26.2  & 16 \\
         \$/1M output token & 0.17 & 0.15 & 0.09 \\
         \bottomrule
    \end{tabular}
    \label{table:per_layer_time}
\end{table}

Under this configuration, the theoretical extreme TPOT for \longcat with SBO can be expressed as:
\[
\begin{aligned}
    \mathrm{TPL}  &= 264 + 236 + 60 + 472 = 1032 ~us, \\[0.6ex]
    \mathrm{TPOT} &= \frac{28 \times \mathrm{TPL}}{1000 \times 1.8} \approx 16~\mathrm{ms},
\end{aligned}
\]
where $\mathrm{TPL}$ denotes the time cost per layer.  

The measured value under batch size 96 is approximately TPOT = 26 ms, which is about 61.5\% of the theoretical value and is on par with DeepSeek-V3 (\textasciitilde 64\%). The gap between measured value and theoretical speed mainly comes from the overhead of small operators and the loss in communication bandwidth.

We apply the same method to calculate the theoretical limits of TPOT and generation cost for DeepSeek-V3 and Qwen3-235B-A22B under TBO scheduling. It can be observed from Table~\ref{table:per_layer_time} that through model system co-design, \longcat achieves significant theoretical improvements in both throughput and latency.

Furthermore, we observed two key insights about \longcat: (1) \longcat exposes not only all-to-all communication and MoE computation, but also an MLA computation. As a result, at the same batch size, \longcat incurs slightly longer per-layer time than DeepSeek-V3. However, due to its significantly reduced layer count, \longcat achieves lower overall latency. (2) \longcat's second MLA is overlapped by the all-to-all combine. This means that in the decoding phase, \longcat can increase the sequence length to a certain extent without substantial latency increase.

\section{Conclusion}

We introduce \longcat, a 560B-parameter MoE model featuring three key innovations: (1) a context-aware dynamical computation mechanism and shortcut-connection MoE, enabling high efficiency in both training and inference, (2) integrated strategies that ensure stable large-scale training, (3) a multi-stage training pipeline that cultivates \longcat's agentic capabilities, allowing it to perform complex tasks requiring iterative reasoning and environmental interaction.
By releasing \longcat as an open-source model, we aim to advance research in efficient MoE architectures, high-quality data strategies, and agentic model development, fostering community-driven innovation in large language models.

\clearpage

\section{Contributions}

The listing of authors is in alphabetical order. Names marked with an asterisk (*) indicate people who have left our team.

\begin{tabular}{p{0.25\textwidth}p{0.25\textwidth}p{0.25\textwidth}p{0.25\textwidth}}
Bayan & Jiahuan Li & Qiyuan Duan & Xuemiao Zhang \\
Bei Li & Jiajun Yang & Ran Meng & Xueyuan Hao \\
Bingye Lei & Jiaming Wang & Rongxiang Weng & Xuezhi Cao \\
Bo Wang & Jian Yang & Ruichen Shao & Xunliang Cai \\
Bolin Rong & Jianchao Tan & Rumei Li & Xurui Yang \\
Chao Wang & Jiaqi Sun & Shizhe Wu & Yan Feng \\
Chao Zhang & Jiaqi Zhang & Shuai Liang & Yang Bai \\
Chen Gao & Jiawei Fu & Shuo Wang & Yang Chen \\
Chen Zhang & Jiawei Yang & Suogui Dang & Yang Yang \\
Cheng Sun & Jiaxi Hu & Tao Fang & Yaqi Huo \\
Chengcheng Han & Jiayu Qin & Tao Li & Yerui Sun \\
Chenguang Xi & Jingang Wang & Tefeng Chen & Yifan Lu \\
Chi Zhang & Jiyuan He & Tianhao Bai & Yifan Zhang \\
Chong Peng & Jun Kuang & Tianhao Zhou & Yipeng Zang \\
Chuan Qin & Junhui Mei & Tingwen Xie & Yitao Zhai \\
Chuyu Zhang & Kai Liang & Wei He & Yiyang Li \\
Cong Chen & Ke He & Wei Huang & Yongjing Yin \\
Congkui Wang & Kefeng Zhang & Wei Liu & Yongkang Lv \\
Dan Ma & Keheng Wang & Wei Shi & Yongwei Zhou \\
Daoru Pan & Keqing He* & Wei Wang & Yu Yang \\
Defei Bu & Liang Gao & Wei Wu & Yuchen Xie \\
Dengchang Zhao & Liang Shi & Weikang Zhao & Yueqing Sun \\
Deyang Kong & Lianhui Ma & Wen Zan & Yuewen Zheng \\
Dishan Liu & Lin Qiu & Wenjie Shi & Yuhuai Wei \\
Feiye Huo & Lingbin Kong & Xi Nan & Yulei Qian \\
Fengcun Li & Lingtong Si & Xi Su & Yunfan Liang \\
Fubao Zhang & Linkun Lyu & Xiang Li & Yunfang Tai \\
Gan Dong & Linsen Guo & Xiang Mei & Yunke Zhao \\
Gang Liu & Liqi Yang & Xiangyang Ji & Zeyang Yu \\
Gang Xu & Lizhi Yan & Xiangyu Xi & Zhao Zhang \\
Ge Li & Mai Xia & Xiangzhou Huang & Zhaohua Yang \\
Guoqiang Tan & Man Gao & Xianpeng Li & Zhenchao Zhang \\
Guoyuan Lin & Manyuan Zhang & Xiao Fu & Zhikang Xia \\
Haihang Jing & Meng Zhou & Xiao Liu & Zhiye Zou \\
Haomin Fu & Mengxia Shen & Xiao Wei & Zhizhao Zeng \\
Haonan Yan & Mingxiang Tuo & Xiaodong Cai & Zhongda Su \\
Haoxing Wen & Mingyang Zhu & Xiaolong Chen & Zhuofan Chen \\
Haozhe Zhao & Peiguang Li & Xiaoqing Liu & Zijian Zhang \\
Hong Liu & Peng Pei & Xiaotong Li & Ziwen Wang \\
Hongmei Shi* & Peng Zhao & Xiaowei Shi & Zixu Jiang \\
Hongyan Hao & Pengcheng Jia & Xiaoyu Li & Zizhe Zhao \\
Hongyin Tang & Pingwei Sun & Xili Wang & Zongyu Wang \\
Huantian Lv & Qi Gu & Xin Chen & Zunhai Su* \\
Hui Su & Qianyun Li & Xing Hu &  LongCat-Flash\\
Jiacheng Li & Qingyuan Li* & Xingyu Miao &  \\
Jiahao Liu & Qiong Huang & Xinyan He &  \\
\end{tabular}

\bibliographystyle{unsrtnat}
\bibliography{references}  

\clearpage

\appendix
\section{Appendix}\label{appendix}

\subsection{Statistics and Case Studies of Dynamic Routing}\label{appendix:pretrain_dynamics}

\begin{figure}[htbp]
\centerline{\includegraphics[width=0.6\columnwidth]{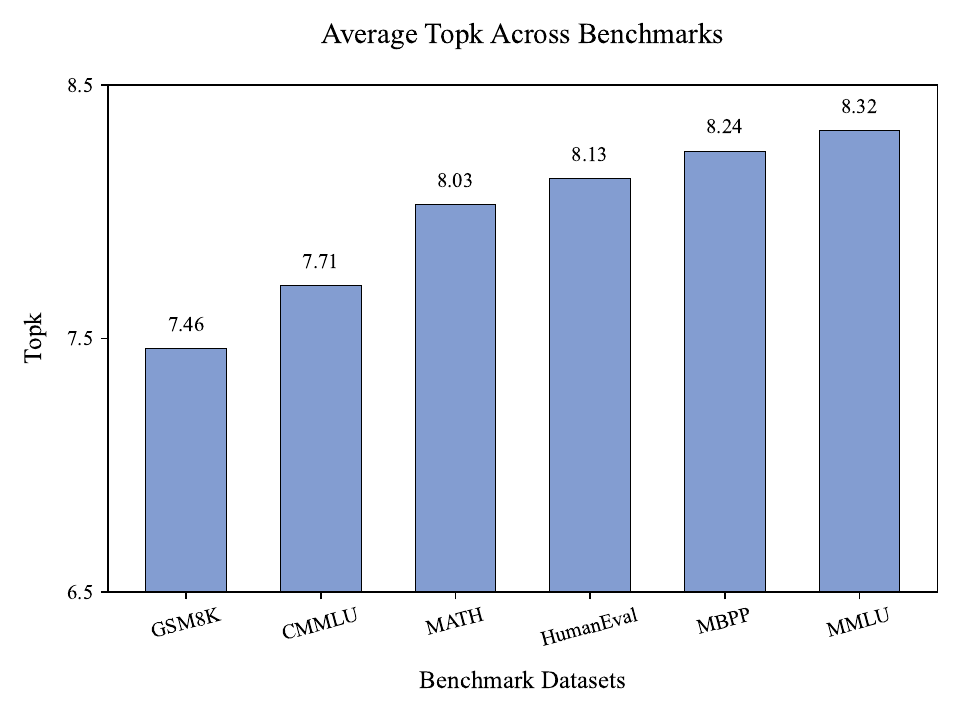}}
\caption{The average number of activated FFN experts across different benchmarks.}
\label{fig:topk-benchmark}
\end{figure}

Figure~\ref{fig:topk-benchmark} shows the average activated FFN experts of \longcat base model across benchmarks. A consistent computational bias favors English tokens over Chinese and mathematical ones.
We present a more detailed expert selection across different layers for several cases in Table~\ref{tab:topk-case}. These cases reveal different patterns of expert selection across layers. In the first layer, function words (including articles, conjunctions, prepositions), numbers and punctuation marks consistently receive lower computational resources. In contrast, the final layer (Layer 28) exhibits less specialized feature allocation compared to Layer 1, though identifiable patterns still exist. For example, in the Chinese text case, tokens preceding punctuation marks tend to be assigned fewer computational resources. We hypothesize that shallow layers prioritize token-internal semantics for allocation, while deeper layers dynamically adjust resources based on predictive complexity, potentially reflecting a hierarchical transition from local feature processing to global prediction optimization.

\begin{table}[htbp]
\centering
\begin{tabular}{|c|}
\hline
\makecell[c]{\vspace{2ex}\small Layer 1 - English \\[-1.5ex]} \\
\includegraphics[width=0.95\linewidth]{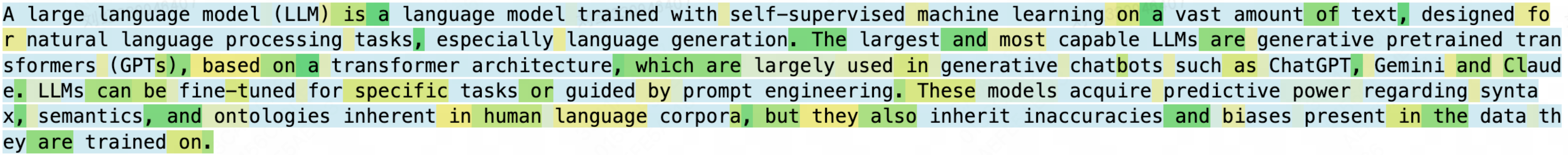} \\ \hline
\makecell[c]{\vspace{2ex}\small Layer 1 - Math \\[-1.5ex]} \\
\includegraphics[width=0.95\linewidth]{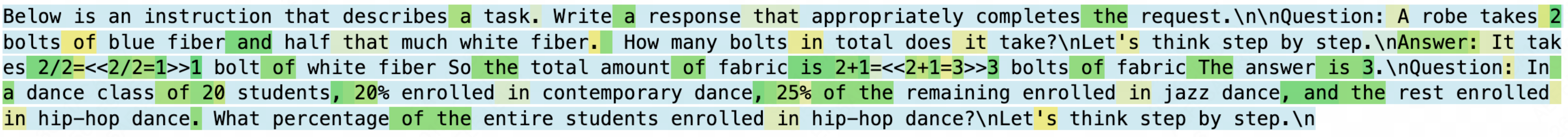} \\ \hline
\makecell[c]{\vspace{2ex}\small Layer 1 - Code \\[-1.5ex]} \\
\includegraphics[width=0.95\linewidth]{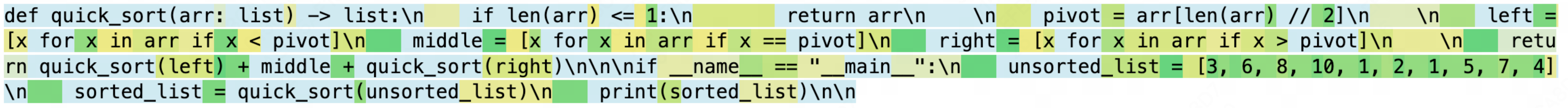} \\ \hline
\makecell[c]{\vspace{2ex}\small Layer 1 - Chinese \\[-1.5ex]} \\
\includegraphics[width=0.95\linewidth]{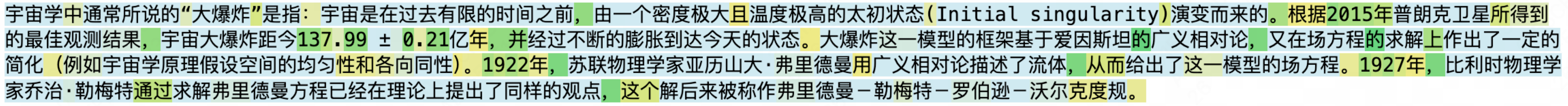} \\ \hline
\makecell[c]{\vspace{2ex}\small Layer 28 - English \\[-1.5ex]} \\
\includegraphics[width=0.95\linewidth]{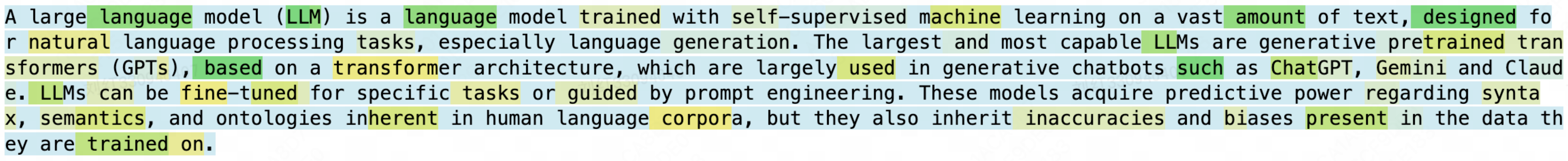} \\ \hline
\makecell[c]{\vspace{2ex}\small Layer 28 - Math \\[-1.5ex]} \\
\includegraphics[width=0.95\linewidth]{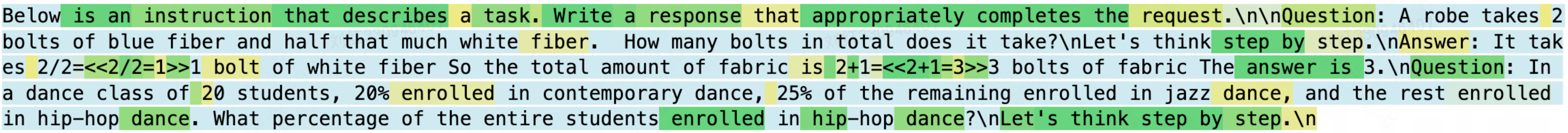} \\ \hline
\makecell[c]{\vspace{2ex}\small Layer 28 - Code \\[-1.5ex]} \\
\includegraphics[width=0.95\linewidth]{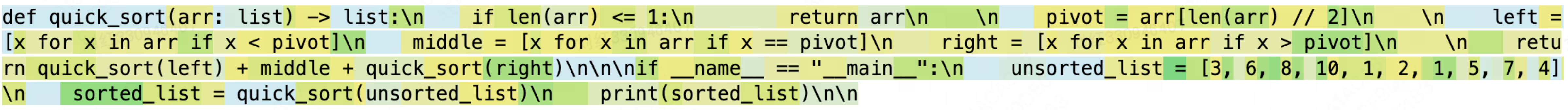} \\ \hline
\makecell[c]{\vspace{2ex}\small Layer 28 - Chinese \\[-1.5ex]} \\
\includegraphics[width=0.95\linewidth]{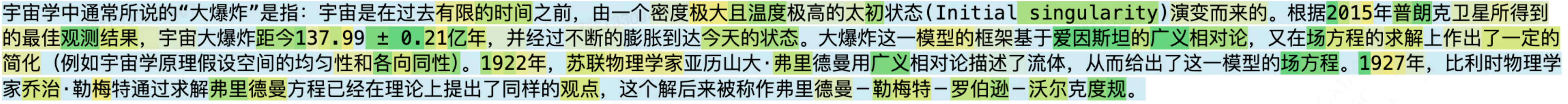} \\ \hline
\end{tabular}
\includegraphics[width=0.5\linewidth]{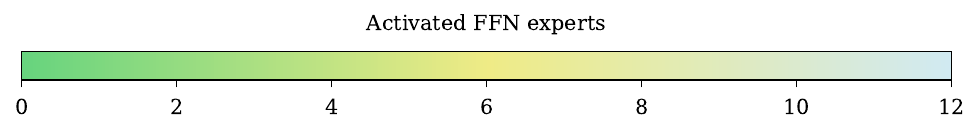}
\caption{The number of activated FFN experts per token across layers.}
\label{tab:topk-case}
\end{table}

\end{document}

%% file: base_model_evaluation.tex
\newcommand{\red}[1]{\textcolor{red}{#1}}

This section presents a comprehensive evaluation of the \longcat base model, including the methodology and results.

\subsubsection{Evaluation Benchmarks and Configurations}
The model evaluation covers four core capabilities: general tasks, general reasoning, mathematical reasoning, and coding. The benchmarks used for assessment include:

\begin{itemize}
    \item \textbf{General Tasks:} MMLU~\citep{hendrycks2021measuringmassivemultitasklanguage}, MMLU-Pro~\citep{wang2024mmluprorobustchallengingmultitask}, C-Eval~\citep{huang2023ceval}, and CMMLU~\citep{li2023cmmlu}.

    \item \textbf{Reasoning Tasks:} GPQA~\citep{rein2023gpqagraduatelevelgoogleproofqa}, SuperGPQA~\citep{pteam2025supergpqascalingllmevaluation}, BBH~\citep{BBH}, PIQA~\citep{piqa}, DROP~\citep{drop}, CLUEWSC~\citep{clue}, and WinoGrande~\citep{winogrande}.
    
    \item \textbf{Math Tasks:} GSM8K~\citep{cobbe2021gsm8k}, MATH~\citep{mathbenchmark}.

    \item \textbf{Coding Tasks:} MBPP+~\citep{humanevalmbppplus}, HumanEval+~\citep{humanevalmbppplus}, MultiPL-E~\citep{cassano2022multiplescalableextensibleapproach}, and CRUXEval~\citep{gu2024cruxevalbenchmarkcodereasoning}. 
\end{itemize}

We compare the \longcat base model with state-of-the-art open-source base MoE models, including DeepSeek-V3.1 Base \citep{deepseekai2025deepseekv3technicalreport}, Llama-4-Maverick Base \citep{meta2025llama4}, and Kimi-K2 Base \citep{Kimi_K2_web_doc}.

To ensure fairness, all models are evaluated under identical pipelines and configurations. For minority results that cannot be reproduced, we directly adopt metrics from public reports and explicitly annotate them in Table~\ref{tab:base_model_results}. The evaluation settings are as follows:

\begin{itemize}
\item General/reasoning/math tasks: Use few-shot prompts to guide output format. Performance is measured via accuracy or F1 score.

\item HumanEval+ and MBPP+: Follow OpenAI's recommended setting \citep{chen2021codex}.

\item MultiPL-E: Follow BigCode Evaluation Harness\citep{bigcode-evaluation-harness}.
\item CRUXEval: Follow the official configuration\footnote{\url{https://github.com/facebookresearch/cruxeval}}, employing 2-shots examples.

\end{itemize}

\subsubsection{Evaluation Results}

Table~\ref{tab:base_model_results} presents the evaluation results across diverse benchmarks. \longcat Base model achieves performance on par with state-of-the-art base models despite its compact active/total parameter size. Although Llama-4-Maverick has fewer activated and total parameters, \longcat Base surpasses both on nearly all benchmarks.

A comparative analysis reveals that \longcat Base matches DeepSeek-V3.1 Base's performance across all domains despite containing fewer parameters. While the two models perform similarly in general tasks, \longcat Base demonstrates a notably advantage on the MMLU-Pro benchmark (featuring challenging questions). For reasoning tasks, \longcat Base attains a higher average score. In math and coding tasks, it outperforms DeepSeek-V3.1 Base on most benchmarks, with only marginal performance gaps observed on CRUXEval and MultiPL-E. Against Kimi K2 Base, \longcat Base shows modestly lower performance in general tasks but achieves parity or superiority in reasoning, math, and coding tasks. 

These results collectively underscore \longcat Base's parameter efficiency, as it delivers competitive or superior performance to larger models across the majority of evaluated benchmarks.

\begin{table}[htbp]
    \caption{Comparison between \longcat and other base models. Values marked with * are sourced from public reports.}
    \centering
    \resizebox{0.8\textwidth}{!}{
    \begin{tabular}{l|ccc|cc}
        \toprule
        Benchmark & \begin{tabular}[c]{@{}c@{}}DeepSeek-V3.1 \\ Base\end{tabular} & \begin{tabular}[c]{@{}c@{}}Llama-4-Maverick \\ Base\end{tabular} & \begin{tabular}[c]{@{}c@{}}Kimi-K2 \\ Base\end{tabular}  & \begin{tabular}[c]{@{}c@{}}\longcat \\ Base\end{tabular} \\ 
        \midrule
        Architecture & MoE & MoE  & MoE  & MoE  \\
        \# Total Params & 671B & 402B & 1043B  & 560B \\
        \# Activated Params & 37B & 17B & 32B  & 27B \\
        \midrule
        \multicolumn{5}{c}{\textbf{General Domains}} \\
        \midrule
        MMLU $_{\text{(acc)}}$ & 87.46 & 84.41 & 87.47  & 87.05 \\
        MMLU-Pro $_{\text{(acc)}}$ & 59.29 & 63.90  & 68.36  & 70.32 \\
        CEval $_{\text{(acc)}}$ & 89.33 & 81.93  & 91.24  & 87.73 \\
        CMMLU $_{\text{(acc)}}$ & 88.21 & 80.71 & 90.35  & 87.19 \\
        \midrule
        \multicolumn{5}{c}{\textbf{General Reasoning}} \\
        \midrule
        GPQA $_{\text{(acc)}}$  & 47.16 & 48.08 & 45.89  & 51.09 \\
        SuperGPQA $_{\text{(acc)}}$ & - & 40.58*  & 44.70*  & 52.03 \\
        BBH $_{\text{(acc)}}$ & 89.46 &87.56 & 89.19 & 90.54  \\
        DROP $_{\text{(f1)}}$  & 80.74 & 77.44 & 69.81  & 78.39  \\
        PIQA $_{\text{(acc)}}$ & 93.00 & 90.59 & 95.10  & 92.33 \\
        WinoGrande $_{\text{(acc)}}$ & 83.50  & 73.32 & 82.87  & 85.08 \\
        CLUEWSC $_{\text{(acc)}}$ & 88.16 & 88.00 & 76.32  & 91.12 \\

        \midrule
        \multicolumn{5}{c}{\textbf{Mathematical Reasoning}} \\
        \midrule

        GSM8K $_{\text{(acc)}}$ & 92.22 & 84.61 & 92.27  & 93.86 \\
        MATH $_{\text{(acc)}}$ & 65.38 & 63.34 & 66.74 & 69.28 \\
        \midrule
        \multicolumn{5}{c}{\textbf{Coding}} \\
        \midrule
        MBPP+ $_{\text{(pass@1)}}$ & 59.26 & 70.11 & 80.49  & 77.25 \\
        HumanEval+ $_{\text{(pass@1)}}$  & 67.07 & 60.37  & 69.84  & 65.85 \\
        MultiPL-E $_{\text{(pass@1)}}$ & 62.00 & 58.35 & 59.22  & 69.25 \\
        CRUXEval-I $_{\text{(pass@1)}}$ & 65.87 & 62.00 & 65.87  & 71.63  \\
        CRUXEval-O $_{\text{(pass@1)}}$ & 71.25 & 64.25 & 68.75  & 75.88 \\
        \bottomrule
    \end{tabular}
    }
    
    \label{tab:base_model_results}
\end{table}

%% file: chat_model_evaluation.tex
We conduct a comprehensive and rigorous evaluation of \longcat after post-training. Specifically, we assess its capabilities across multiple dimensions, including general domains, instruction following, mathematical reasoning, general reasoning, coding, agentic tool use, and safty tasks.

\subsubsection{Evaluation Benchmarks and Configurations}

The evaluation employs the following benchmarks:
\begin{itemize}
    \item \textbf{General Domains:} MMLU~\citep{hendrycks2021measuringmassivemultitasklanguage}, MMLU-Pro~\citep{wang2024mmluprorobustchallengingmultitask}, 
    ArenaHard~\citep{li2024crowdsourced, arenahard2024}, CEval~\citep{huang2023ceval}, and CMMLU~\citep{li2023cmmlu}.
    \item \textbf{Instruction Following:} IFEval~\citep{zhou2023ifeval}, COLLIE~\citep{yao2024collie}, and Meeseeks~\citep{meeseeks}, Meeseeks evaluates models' instruction-following capabilities in multi-turn scenarios through an iterative feedback framework that simulates realistic human-LLM interactions, enabling models to self-correct based on turn-specific failures and better reflect real-world usage patterns.
    \item \textbf{Mathematical Reasoning:}  MATH500~\citep{math500}, AIME24~\citep{AIME24}, AIME25~\citep{AIME25}, and BeyondAIME~\citep{bytedance_seed_2025_beyondaime}.
    \item \textbf{General Reasoning:}  GPQA-diamond~\citep{rein2023gpqagraduatelevelgoogleproofqa}, DROP~\citep{drop}, ZebraLogic~\citep{lin2025zebralogic}, and GraphWalks~\citep{graphwalks}.
    \item \textbf{Coding:}  Humaneval+~\citep{humanevalmbppplus}, MBPP+~\citep{humanevalmbppplus}, 
    LiveCodeBench (2024.08-2025.05)~\citep{jain2025livecodebench}, 
    SWE-Bench-Verified~\citep{jimenez2024swebench}, and TerminalBench~\citep{tbench_2025}.
    \item \textbf{Agentic Tool Use:}  $\tau^2$-Bench~\citep{tau2-bench} and AceBench~\citep{chen2025acebench}. Furthermore, we develop a high-quality proprietary benchmark, VitaBench, leveraging Meituan's comprehensive real-world business scenarios to systematically evaluate models' capabilities in addressing complex real-world tasks. Within VitaBench, to comprehensively assess models' generalized agentic capabilities, we deliberately curate cross-domain quotidian scenarios and explicitly delineate inter-tool dependencies, eschewing the provision of extensive domain-specific policies. Our benchmark emphasizes three critical dimensions of complexity: tool set complexity (characterized by dense tool graphs averaging over 30 available tools per task), reasoning complexity, and user interaction complexity (featuring challenging user personas with an average exceeding 60 interaction rounds per task for evaluated models). The complete benchmark dataset, along with detailed construction methodologies and comprehensive result analysis, will be fully released in subsequent work.
\end{itemize}

We also evaluate the safety performance of \longcat. Specifically, we conduct evaluations on four major risk categories:
\begin{itemize}
    \item \textbf{Harmful}: Violence, hate Speech, insulting, harassment and bullying, self-harm and suicide, adult content, etc. 
    \item \textbf{Criminal}: Illegal activities, underage violations, extreme terrorism and violence, etc. 
    \item \textbf{Misinformation}: misinformation and disinformation, unsafe practices, hallucination, etc. 
    \item \textbf{Privacy}: privacy violation, infringement, etc.
\end{itemize}

Within each category, a sufficient number of private test queries are constructed, followed by a comprehensive manual review to ensure the accuracy of their classification and the reliability of their quality.

We compare the chat version of \longcat with several contemporary non-thinking chat models, including DeepSeek-V3.1~\citep{deepseekai2025deepseekv3technicalreport}, Qwen3-235B-A22B (2507 version)~\citep{yang2025qwen3}, Kimi-K2~\citep{Kimi_K2_web_doc}, GPT-4.1~\citep{gpt4.1}, Claude4-Sonnet~\citep{claude4}, and Gemini2.5-Flash~\citep{comanici2025gemini}. For closed-source models, we conduct evaluations through their official APIs. For models supporting both thinking and non-thinking modes (Qwen3-235B-A22B, Gemini2.5-Flash, and Claude4-Sonnet), we explicitly configure these models to operate in non-thinking mode for a fair comparison.

For each benchmark category, we employ the following specialized metrics and settings:
\begin{itemize}
    \item \textbf{General domain benchmarks:} We use accuracy as the evaluation metric, except for ArenaHard-V2. Unlike the original benchmarks that rely on exact-match (EM) for correctness judgment, we employ a scoring model to assess whether model responses align with reference answers. Since our scoring model recognizes semantically correct answers even without exact textual matches, reported values may be slightly higher than originally documented.
    \item \textbf{Instruction following benchmarks:} We design regular expressions based on instruction rules to verify compliance. Rule-based and model-based answer span extraction tools are additionally employed to support this evaluation.
    \item \textbf{Mathematical reasoning benchmarks:} We apply the aforementioned scoring model for MATH500, and the averaged EM scores over $10$ runs for AIME-related benchmarks.
    \item \textbf{General reasoning benchmarks:}  We apply the scoring model for GPQA-diamond, calculate the F1 score for DROP, adopt rule-based matching for ZebraLogic, and use the precision metric for GraphWalk following the official implementation on its 128k context length subset.
    \item \textbf{Coding benchmarks:} Each problem is scored 1 if the model's response passes all test cases in a sandbox environment or matches a specific state, otherwise 0. The final score is the average across all problems. We adopt the script provided by OpenAI\footnote{\url{https://github.com/bigcode-project/bigcode-evaluation-harness}} to evaluate Humaneval+ and MBPP+, and the official scripts for the others. Specifically, for SWE-Bench-Verified, we use R2E-Gym\footnote{\url{https://github.com/R2E-Gym/R2E-Gym}} (Openhands scraffold) with runs limited to 100 iterations to evaluate all models except Deepseek-V3.1 and Claude4-Sonnet; their results are sourced from other public reports.
    \item \textbf{Agentic tool use benchmarks:} We utilize official benchmark frameworks to ensure fairness and reproducibility. For AceBench, we use direct prompting rather than function calling. For our proposed \textbf{VitaBench}, given the inherent long-context characteristics of agentic tasks, we employ a sliding window mechanism to systematically evaluate task completion status throughout the entire execution trajectory, facilitating continuous updates to the completion status of individual checklist components.
\end{itemize}

\subsubsection{Evaluation Results}

As detailed in Table \ref{tab:chat_model_eval_results}, our comprehensive evaluation reveals that \longcat is a powerful and versatile model. It consistently demonstrates leading performance in different domains, often outperforming contemporary models across a wide array of challenging tasks with relatively fewer activated parameters. The following analysis provides a detailed breakdown of its impressive capabilities across different dimensions.

\textbf{General Domains}  In general domain knowledge, \longcat demonstrates a strong and well-rounded performance. It achieves an excellent score of 86.50 on ArenaHard-V2, ranking second among all evaluated models and showcasing its robust capabilities in challenging head-to-head comparisons. On foundational benchmarks, it remains highly competitive, scoring 89.71 on MMLU and 90.44 on CEval. These results are comparable to leading models, and notably, are achieved with fewer parameters than competitors like DeepSeek-V3.1 and Kimi-K2, indicating high efficiency.

\textbf{Instruction Following}  \longcat exhibits state-of-the-art instruction following capabilities. It achieves the highest score of 89.65 on IFEval, outperforming all other models and demonstrating superior reliability in adhering to complex and nuanced directives. Furthermore, it secures the best score on COLLIE (57.10) and Meeseeks-zh (43.03), underscoring its exceptional proficiency across diverse and challenging instruction sets in both English and Chinese.

\textbf{Mathematical Reasoning}  In mathematical reasoning, \longcat shows powerful and advanced capabilities. While its score on MATH500 (96.40) is highly competent, its strength is particularly evident in more complex, competition-level benchmarks. It delivers excellent, top-tier results on AIME25 (61.25) and BeyondAIME (43.00), ranking among the best-performing models in these challenging domains. This highlights its advanced capacity for sophisticated, multi-step logical deduction and problem-solving.

\textbf{General Reasoning}  For general reasoning tasks, \longcat's performance is also solid. It demonstrates exceptional strength in structured logical deduction, achieving a score of 89.30 on ZebraLogic, which is among the top competitors. It also obtains a competitive score of 79.06 on the reading comprehension benchmark DROP. Conversely, its results on GPQA-diamond (73.23) and GraphWalks (51.05) indicate an opportunity for further improvement, particularly in enhancing its capabilities for analyzing structured data within extremely long contexts.

\textbf{Coding}  \longcat displays a promising and capable profile in the coding domain. Its standout performance is on TerminalBench, where it achieves a score of 39.51, ranking second and demonstrating excellent proficiency in practical, agentic command-line tasks. It is also competitive on the SWE-Bench-Verified benchmark with a score of 60.4. On foundational code generation tasks such as Humaneval+ and MBPP+, its performance is solid, yet there remains potential for future optimization to align with the leading models.

\textbf{Agentic Tool Use}  \longcat demonstrates a clear advantage in using agentic tool use domain, notably outperforming other models on $\tau^2$-Bench even when compared to models with more parameters. In highly complex scenarios, it achieves the highest score of 24.30 on VitaBench, demonstrated strong capability in complex scenarios.

\textbf{Safety}  \longcat showed outstanding capability in identifying and mitigating risks on the whole, particularly in the domains of Harmful and Criminal compared to other models.

\begin{table*}[htbp] 
     \centering 
     \caption{Evaluation results of frontier chat models. Values marked with * are sourced from other public reports. Note that DeepSeek-V3.1, Qwen3-235B-A22B, Gemini2.5-Flash, and Claude4-Sonnet are evaluated under their non-thinking mode.}
     \label{tab:chat_model_eval_results} 
     \resizebox{\textwidth}{!}{ 
     \begin{tabular}{l|ccc|ccc|c} 
         \toprule 
         \textbf{Benchmark} & \begin{tabular}[c]{@{}c@{}}\textbf{DeepSeek} \\ \textbf{V3.1}\end{tabular} & \begin{tabular}[c]{@{}c@{}}\textbf{Qwen3} \\ \textbf{MoE-2507}\end{tabular} & \begin{tabular}[c]{@{}c@{}}\textbf{Kimi-K2} \\ \end{tabular} & \textbf{GPT-4.1} & \begin{tabular}[c]{@{}c@{}}\textbf{Claude4} \\ \textbf{Sonnet}\end{tabular} & \begin{tabular}[c]{@{}c@{}}\textbf{Gemini2.5} \\ \textbf{Flash}\end{tabular} & \begin{tabular}[c]{@{}c@{}}\textbf{\longcat} \\ \end{tabular} \\ 
         \midrule 
         
Architecture & MoE & MoE & MoE & - & - & - & MoE \\ 
         \# Total Params & 671B & 235B & 1043B & - & - & - & 560B \\ 
         \# Activated Params & 37B & 22B & 32B & - & - & - & 27B \\ 
         \midrule 
         \multicolumn{8}{c}{\textbf{General Domains}} \\ 
  
        \midrule 
         MMLU $_{\text{(acc)}}$ & 90.96 & 90.23 & 89.86 & 89.64 & 91.75 & 86.33 & 89.71 \\ 
         MMLU-Pro $_{\text{(acc)}}$  & 84.45 & 84.83 & 82.06 & 81.72 & 83.74 & 81.95 & 82.68 \\ 
         ArenaHard-V2 & 84.10 & 88.20 & 85.70  & 61.50 & 62.10 & 77.00 & 86.50 \\ 
         CEval $_{\text{(acc)}}$ & 89.21 & 92.70 & 91.26 & 79.53 & 86.63 & 78.78 & 90.44 \\ 
         CMMLU $_{\text{(acc)}}$ & 88.04 & 88.14 & 89.66 & 77.65 & 86.51 & 78.30 & 84.34 \\ 
         \midrule 
         \multicolumn{8}{c}{\textbf{Instruction Following}} \\ 
        
 \midrule 
         IFEval $_{\text{(acc)}}$ & 86.69 & 88.54 & 88.91 & 85.58 & 88.35 & 83.92 & 89.65 \\ 
         COLLIE $_{\text{(acc)}}$ & 43.80 & 49.71 & 56.34 & 50.00 & 51.22 & 48.60 & 57.10 \\ 
         Meeseeks-zh $_{\text{(acc)}}$ & 33.83 & 35.32 & 42.79 & 41.54 & 35.07 & 34.84 & 43.03 \\ 
         \midrule 
       
  \multicolumn{8}{c}{\textbf{Mathematical Reasoning}} \\ 
         \midrule 
         MATH500 $_{\text{(acc)}}$  & 96.08 & 98.80 & 97.60 & 90.60 & 93.80 & 98.40 & 96.40 \\ 
         AIME24 $_{\text{(avg@10)}}$  & 66.30* & 81.67 & 69.60* & 47.00 & 47.00 & 79.67 &  70.42 \\ 
         AIME25 $_{\text{(avg@10)}}$  & 49.27 & 68.33 & 50.66 & 32.00 & 37.00 & 67.33 & 61.25 \\ 
    
     BeyondAIME $_{\text{(avg@10)}}$ & 36.50 & 57.60 & 36.60 & 22.10 & 20.50 & 44.20 & 43.00 \\ 
         \midrule 
         \multicolumn{8}{c}{\textbf{General Reasoning}} \\ 
         \midrule 
         GPQA-diamond $_{\text{(acc)}}$ & 74.90* & 77.43 & 75.76 & 67.68 & 70.71 & 80.30 & 73.23 \\ 
         DROP $_{\text{(f1)}}$  & 84.19 & 78.57 & 89.04 & 66.94 & 73.06 & 45.03 & 79.06 \\ 
         ZebraLogic $_{\text{(acc)}}$  & 85.30 & 94.22 & 89.11 & 56.30* & 80.10 & 57.00 & 89.30 \\ 
         GraphWalks-128k $_{\text{(precision)}}$  & 73.54 & 80.72 & 47.50 & 85.02 & 80.57 & 64.83 & 51.05 \\ 
         \midrule 
         \multicolumn{8}{c}{\textbf{Coding}} \\ 
         \midrule 
   
         LiveCodeBench $_{\text{(pass@1)}}$ & 56.40*  & 46.48 & 46.70 & 39.21 & 45.59 & 39.65 & 48.02 \\ 
      Humaneval+ $_{\text{(pass@1)}}$  & 92.68 & 94.51 & 85.98 & 93.29 & 94.51 & 87.80 & 88.41 \\ 
         MBPP+ $_{\text{(pass@1)}}$  & 79.89 & 79.89 & 81.75 & 79.37 & 80.16 & 76.19 & 79.63 \\ 
         SWE-Bench-Verified $_{\text{(acc)}}$  & 66.00* & 48.40 & 64.60 & 51.00 & 68.00* & 42.40 & 60.40 \\ 
         TerminalBench $_{\text{(acc)}}$ & 31.30* & 17.28 & 25.93 & 28.40 & 40.74 & 12.35 & 39.51 \\ 
         \midrule 
         \multicolumn{8}{c}{\textbf{Agentic Tool Use}} \\ 
         \midrule 
         
         $\tau^2$-Bench (telecom) $_{\text{(avg@4)}}$  & 38.50 & 25.60 & 67.50 & 35.20 & 46.20 & 16.50 & 73.68 \\ 
         $\tau^2$-Bench (airline) $_{\text{(avg@4)}}$ & 46.00 & 48.00 & 54.20 & 56.00 & 60.00 & 41.50 & 58.00 \\ 
         $\tau^2$-Bench (retail) $_{\text{(avg@4)}}$  & 64.90 & 70.50 & 70.80 & 74.10 & 80.00 & 64.80 & 71.27 \\ 
         AceBench $_{\text{(acc)}}$  & 71.70 & 76.00 & 71.70 & 80.10* & 76.10 & 74.50* & 76.10 \\
         VitaBench $_{\text{(avg@4)}}$  & 20.30 & 8.50 & 18.20 & 19.00 & 23.00 & 8.00 & 24.30 \\
         \midrule 
         \multicolumn{8}{c}{\textbf{Safety}} \\ 
         \midrule 
         Harmful        & 82.79 & 80.82 & 53.91 & 56.19 &  66.56  & - & 83.98   \\ 
        Criminal       & 87.83 & 89.13 & 77.19 & 81.58 & 87.58  & - & 91.24   \\
        Misinformation & 83.17 & 77.76 & 42.68 & 45.49 & 54.91  & - & 81.72   \\
        Privacy        & 98.80 & 98.80 & 96.39 & 98.80 & 100.00 & - & 93.98   \\
         \bottomrule 
     \end{tabular}
     } 
 \end{table*}